\documentclass{article}

\usepackage[preprint]{neurips_2023}

\usepackage[utf8]{inputenc} % allow utf-8 input
\usepackage[T1]{fontenc}    % use 8-bit T1 fonts
\usepackage{hyperref}       % hyperlinks
\usepackage{url}            % simple URL typesetting
\usepackage{booktabs}       % professional-quality tables
\usepackage{amsfonts}       % blackboard math symbols
\usepackage{nicefrac}       % compact symbols for 1/2, etc.
\usepackage{microtype}      % microtypography
\usepackage{xcolor}         % colors
\usepackage{amsmath}
\usepackage{multirow}
\usepackage{graphicx}
\usepackage{gensymb}
\usepackage{courier}
\usepackage{natbib}
\usepackage{textcomp}
\usepackage{pifont}
\usepackage{parcolumns}
\usepackage[most]{tcolorbox}

\definecolor{custombackground}{HTML}{F0FFFF} % Light gray background
\definecolor{customframe}{HTML}{2E8B57}      % Darker gray frame
\definecolor{customtitle}{HTML}{F0FFFF}      % Almost black title 

\newcommand{\tabincell}[2]{\begin{tabular}{@{}#1@{}}#2\end{tabular}}
\newtcolorbox{dialoguebox}[1][]{
    colback=custombackground, % Use the custom background color
    colframe=customframe,     % Use the custom frame color
    coltitle=customtitle,     % Use the custom title color
    title=#1,
    fonttitle=\bfseries,
    sharp corners,
    boxsep=5pt,
    left=5pt,
    right=5pt,
    before skip=10pt,
    after skip=10pt,
    overlay unbroken and first={ 
      \node[anchor=north east,inner sep=2pt] 
      at (frame.north east) {\textbf{Dialogue}};
    }
}

\setcitestyle{numbers,square}

\title{Astrocyte-Enabled Advancements in Spiking Neural Networks for Large Language Modeling}

\author{ 
  {Guobin Shen$^{1, 2, 4}$\footnotemark[1], \ Dongcheng Zhao$^{1, 2}$\footnotemark[1], \ Yiting Dong$^{1, 2, 4}$,} \\ 
  \textbf{\ Yang Li$^{1, 2, 5}$, \ Jindong Li$^{1, 2, 5}$, \ Kang Sun$^{2}$, \ Yi Zeng$^{1, 2, 3, 4, 5}$\footnotemark[2]}\\
  $^1$ Brain-inspired Cognitive Intelligence Lab, Institute of Automation, Chinese Academy of Sciences\\
  $^2$ Center for Long-term Artificial Intelligence \\
  $^3$ Center for Excellence in Brain Science and Intelligence Technology, CAS\\
  $^4$ School of Future Technology, University of Chinese Academy of Sciences \\
  $^5$ School of Artificial Intelligence, University of Chinese Academy of Sciences \\
  \texttt{\{shenguobin2021, zhaodongcheng2016, dongyiting2020,} \\
  \texttt{liyang2019, lijindong2022\}@ia.ac.cn,} \\ 
  \texttt{kang.sun@long-term-ai.center, yi.zeng@ia.ac.cn}
}

\begin{document}

\maketitle

\renewcommand{\thefootnote}{\fnsymbol{footnote}}
\footnotetext[1]{Equal contribution.}
\footnotetext[2]{Corresponding Author.}
\renewcommand{\thefootnote}{\arabic{footnote}}

\begin{abstract}
    % 1. 脉冲神经网络是生物上可行的高性能神经计算模型. 2. 脉冲神经网络大多使用了LIF神经元模型, 限制了其表现力和动态多样性. 3. 神经胶质细胞占了脑细胞总数的50%-90%, 具有重要的发育, 结构和代谢功能. 4. 最近的实验也表明, 星形胶质细胞直接参与了学习和记忆等核心的认知过程. 5. 受到人工智能和生物学证据的启发我们设计了星形胶质细胞调控的脉冲神经网络模型. 6. 我们设计的这种模型为神经元-胶质细胞提供了一个具体的计算范式, 并具有良好的拓展性, 能够切换不同的计算模式, 以适应不同的硬件. 7. 这种生物启发的计算范式具有比传统的Transformer更低的复杂度, 并具有扩展为大规模语言模型的潜力. 8. 星形胶质细胞调控的脉冲神经网络模型展现出了远超于以往的脉冲神经网络的能力, 并可能潜在地解释了大脑神经元-胶质细胞的在认知中的普遍性和强大的计算功能. 

    % 在错综复杂的大脑织锦中，占细胞总数一半以上的神经胶质细胞在发育、结构和新陈代谢中发挥着关键作用。最近的研究发现，星形胶质细胞是胶质细胞的一个亚群，是通过三方突触调节神经活动并直接影响学习和记忆等认知过程的关键角色。尽管如此，传统的尖峰神经网络（SNN）模型仍以神经元为主要特征，而忽略了神经胶质细胞对神经动态的深远影响。在这些生物学见解的指导下，并借鉴人工智能的进步，我们引入了星形胶质细胞调制的尖峰单元。这一创新结构为神经元与神经胶质细胞的相互作用提供了一个切实可行的计算框架，展示了在各种硬件平台上的通用性。在 AM-SU 的基础上，我们提出的星形胶质细胞调制的 SNN 与传统架构相比，复杂性有所降低，这表明它对广泛的语言模型具有潜在的可扩展性。最终，我们的星形胶质细胞调制 SNN 超越了之前的模型，将神经元与胶质细胞的相互作用视为认知计算的基本组成部分。这项工作缩小了生物合理性与神经建模之间的计算鸿沟，证实了将星形胶质细胞动力学整合到智能神经网络中的可行性和优势，为未来包含神经元和胶质细胞实体的生物启发神经计算研究奠定了基础。
    %在大脑这一错综复杂的神经结构中，星形胶质细胞在发育、结构和新陈代谢方面扮演着关键角色。这些细胞通过三方突触机制调控神经活动，并直接影响学习和记忆等认知过程。然而，尽管星形胶质细胞的重要性日益凸显，传统的尖峰神经网络（SNN）模型依旧以神经元为主，忽略了星形胶质细胞对神经动态的深刻影响。受到这些生物学发现的启发，我们开发了一种星形胶质细胞调制的尖峰单元（AM-SU），这一创新架构将神经元与星形胶质细胞的相互作用融入到计算框架中，展现了其在不同硬件平台上的广泛适用性。我们的星形胶质细胞调制脉冲神经网络（AstroSNN）在记忆保持和自然语言生成（NLG）等任务中展现出卓越性能，特别是在处理长期依赖性和复杂语言结构方面表现显著。AstroSNN的设计不仅提高了模型在生物学上的真实性，还引入了创新的计算动力学机制，使得网络能够更有效地处理复杂的时间依赖关系。此外，AstroSNN在实际应用中表现出低延迟、高吞吐量和较小的内存占用，非常适合于资源受限的环境。AstroSNN的提出不仅挑战了当前的神经网络架构范式，也为计算神经科学和人工智能的进一步研究提供了丰富的平台。通过将星形胶质细胞动力学成功整合到智能神经网络中，我们的工作缩小了生物合理性与神经建模之间的差距，为未来包含神经元和星形胶质细胞的生物启发型神经计算研究奠定了基础。
    Within the complex neuroarchitecture of the brain, astrocytes play crucial roles in development, structure, and metabolism. These cells regulate neural activity through tripartite synapses, directly impacting cognitive processes such as learning and memory. Despite the growing recognition of astrocytes' significance, traditional Spiking Neural Network (SNN) models remain predominantly neuron-centric, overlooking the profound influence of astrocytes on neural dynamics. Inspired by these biological insights, we have developed an Astrocyte-Modulated Spiking Unit (AM-SU), an innovative framework that integrates neuron-astrocyte interactions into the computational paradigm, demonstrating wide applicability across various hardware platforms. Our Astrocyte-Modulated Spiking Neural Network (AstroSNN) exhibits exceptional performance in tasks involving memory retention and natural language generation, particularly in handling long-term dependencies and complex linguistic structures. The design of AstroSNN not only enhances its biological authenticity but also introduces novel computational dynamics, enabling more effective processing of complex temporal dependencies. Furthermore, AstroSNN shows low latency, high throughput, and reduced memory usage in practical applications, making it highly suitable for resource-constrained environments.  By successfully integrating astrocytic dynamics into intelligent neural networks, our work narrows the gap between biological plausibility and neural modeling, laying the groundwork for future biologically-inspired neural computing research that includes both neurons and astrocytes.
    % In the intricate tapestry of the brain, glial cells—comprising over half the total cell population—play pivotal roles in development, structure, and metabolism. Recent insights have cast astrocytes, a subset of glia, as key players in modulating neural activity through tripartite synapses and directly influencing cognitive processes such as learning and memory. Despite this, traditional Spiking Neural Network (SNN) models predominantly feature neurons, discounting the profound influence of glial cells on neural dynamics. Guided by these biological insights and informed by advances in artificial intelligence, we introduce an Astrocyte-Modulated Spiking Unit (AM-SU). This innovative construct provides a tangible computational framework for neuron-glial interplay, demonstrating versatility across various hardware platforms. Building upon the AM-SU foundation, the astrocyte-modulated SNN we propose, exhibits reduced complexity when juxtaposed with conventional architectures, suggesting its potential scalability for extensive language models. Ultimately, our Astrocyte-Modulated Spiking Neural Network (AstroSNN) surpasses previous models, positing the neuron-glial interplay as a fundamental component of cognitive computation. This work narrows the computational chasm between biological plausibility and neural modeling, substantiating the viability and advantages of integrating astrocytic dynamics into intelligent neural networks, potentially laying the groundwork for future biologically-inspired neural computing research inclusive of both neuronal and glial entities.
\end{abstract}

% -------------- For refs. --------------

% 星形胶质细胞生物学入门。神经胶质细胞是大脑中除神经元外的另一种主要细胞类型。神经胶质细胞与神经元的确切比例存在争议，但大约在 1:1 到 10:1 之间 (25)。研究最深入的神经胶质细胞类型是星形胶质细胞。星形胶质细胞的一个显著特点是，单个星形胶质细胞会与附近数千至数百万个突触形成连接（26）。例如，单个人类星形胶质细胞可在单个区域内覆盖 27 万至 200 万个突触（27）。星形胶质细胞大多无电，通过细胞内钙离子（Ca2+）的动态变化来编码信息。在大脑的大部分区域，神经元和星形胶质细胞紧密交织在一起。例如，在海马中，多达 60% 的轴突-树突突触被称为突触过程的星形胶质细胞细胞膜包裹（28）。在小脑中，这一数字甚至更高。这种三向排列（突触前轴突、突触后树突、星形胶质细胞突触）非常常见，因此被命名为三方突触（24）。星形胶质细胞过程含有与它们所包裹的突触部位释放的神经递质相对应的受体。例如，基底节的星形胶质细胞对多巴胺敏感，而大脑皮层的星形胶质细胞对谷氨酸敏感（29）。尽管突触前神经递质相同，但突触后神经元和星形胶质细胞的反应却截然不同： 神经元主要通过动作电位编码信息，而星形胶质细胞则通过细胞间游离钙的升高编码信息。重要的是，神经元与星形胶质细胞之间的信号传递可以触发星形胶质细胞对神经元的反向反应，从而在神经细胞和星形胶质细胞之间建立起反馈回路。星形胶质细胞可根据情况抑制或促进突触（30）。例如，据观察，下丘脑中的星形胶质细胞可通过相同的公共因子成倍放大它们所包被的兴奋性突触（31）。有趣的是，大脑中星形胶质细胞与星形胶质细胞之间也存在广泛的交流。星形胶质细胞彼此形成大规模网络（26）。这些网络在空间上呈瓦片状，细胞间距为几十微米（32）。神经元主要通过尖峰信号进行交流，而星形胶质细胞则不同，它们通过在细胞体、突起和端足之间传播的钙波进行交流（33）。这些波的速度为每秒几十微米。人们认为，这些波可以用来同步神经群和协调重要的神经过程（34）。在这大量的生物现象中，以下四点对我们的数学模型非常重要： - 大脑中的大多数突触都是三方的（突触前神经元、突触后神经元、星形胶质细胞过程）。- 星形胶质细胞过程和突触之间存在反馈回路。星形胶质细胞过程会对突触前神经活动做出反应，使细胞内钙离子（Ca2+）升高，进而释放胶质递质来调节突触。这种调节可以是促进性的，也可以是抑制性的。- 神经元→星形胶质细胞的信号通路具有可塑性。- 邻近的星形胶质细胞过程可以在空间上平均其 Ca2+ 水平。

\section{Introduction}
Traditionally, neurons have been regarded as the principal agents of neural information transmission, while the functions of astrocytes have often not received adequate attention. In contemporary neuroscientific research, sustantial advancements have markedly augmented our comprehension of astrocytes' integral role in neural functionalities. Through rigorous investigation of the 'tripartite synapse model'(an intricate framework comprising neurons, astrocytes, and synaptic architectures), the critical contribution of astrocytes to synaptic functionalities is progressively being elucidated. These cells execute precise modulation of neurotransmitters within the synaptic cleft and engage actively in inter-neuronal communication via the secretion of gliotransmitters, thereby exerting a substantial influence on neuropathologies and cognitive processes~\cite{perea2009tripartite,santello2012gliotransmission,halassa2007tripartite,farhy2018astrocytes}. This novel comprehension elucidates that astrocytes are instrumental in the modulation of synaptic plasticity~\cite{hosli2022decoupling,corkrum2020dopamine}, the reorganization of neural circuits~\cite{illes2019astroglia,halassa2010integrated}, and the genesis of memory~\cite{suzuki2011astrocyte,kol2020astrocytes,de2022multiple,becker2022modulation}. By exerting influence on both the release and reabsorption of neurotransmitters, astrocytes directly engage in the modulation and processing of neural signals. These observations imply that the historical underrepresentation of astrocytes in prior neural network models may constitute a significant factor underlying the observed disparities between the functional output of these models and the innate capacities of biological neural networks.

Spiking Neural Networks (SNNs), deriving inspiration from biological paradigms, have emerged as a pivotal area of inquiry within the realms of computational neuroscience and machine learning~\cite{maass1997networks,roy2019towards,zeng2023braincog}. The burgeoning interest in SNNs is attributable to their proficiency in emulating the dynamic and temporal processing traits inherent in biological neural systems. Researchers have conducted extensive studies on various aspects of SNNs to enhance their computational power and efficiency, including neuron models, training algorithms, and network architectures. Concerning neuron modeling, a diversity of adaptive neuron models has been innovatively developed~\cite{fang2021incorporating,shen2023dive,zhao2022backeisnn}. These models are specifically designed to enhance the computational abilities and adaptability of neurons, thereby significantly improving the overall performance of neural networks. In the realm of synaptic plasticity algorithmic research, significant progress has been achieved through the development of methods~\cite{zhang2018highly,shen2023metaplasticity,dong2023unsupervised,shen2022backpropagation}. These advancements have substantially enhanced the proficiency of SNNs in managing complex computational tasks. Furthermore, pioneering studies~\cite{zhou2022spikformer,shen2023brain} have facilitated enhancements in network artectures. These advancements have enabled SNNsto exhibit greater adaptability and versatility in meeting the demands of a diverse array of task-specific requirements. Nevertheless, when tasked with processing extended temporal sequences, especially within the ambit of large-scale language model applications, SNNs demonstrate conspicuous constraints in memory capacity. This impediment predominantly originates from the spiking neurons' limited time-dependent characteristics inherent within SNNs. In addition, contemporary SNNs encounter challenges in temporal processing attributable to coupling issues and a deficiency in parallel processing capabilities, which consequentially curtail their efficiency in both training and deployment in expansive applications. Intriguingly, astrocytes, which is a type of neural cell in the biological brain that forms connections with multiple neuron synapses, introduce an ancillary layer for long-term information processing within neural networks via their interactions with neurons. This astrocyte-facilitated mechanism efficaciously amplifies the memory capacity and enhances the complexity involved in processing time-series data within neural networks.

\begin{figure}
    \centering
    \includegraphics[width=1.0\linewidth]{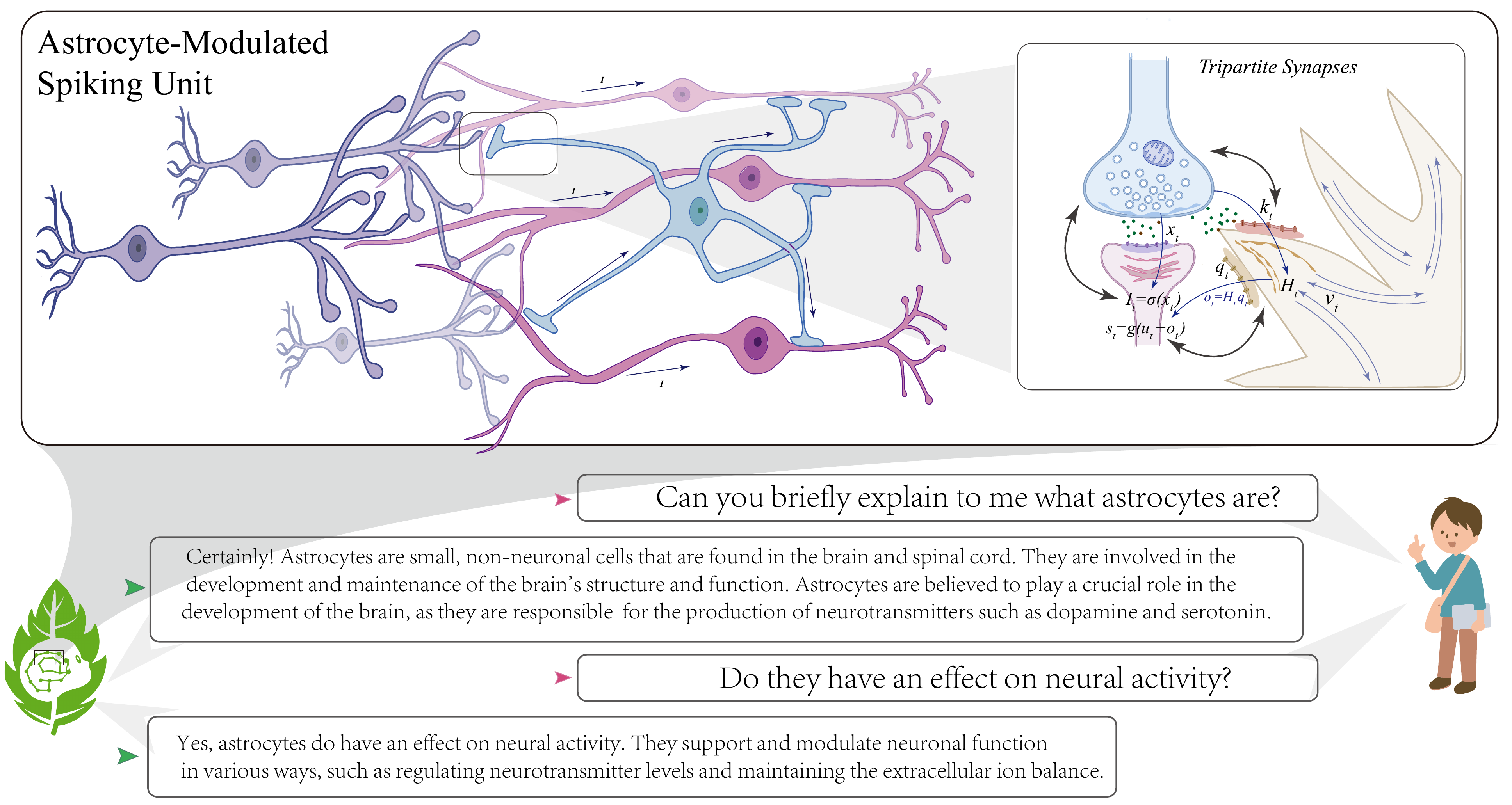}
    % 星形胶质细胞调制尖峰单元（AM-SU）图解。左图：星形胶质细胞与神经元交织在一起，影响神经活动的复合尖峰单元。右图 星形胶质细胞在三方突触中的示意图，模拟突触传递和可塑性，并调节神经活动。下图 星形胶质细胞调控的尖峰神经网络可增强大型语言模型的能力，并产生对话能力。
    \caption{Illustration of Astrocyte-Modulated Spiking Units (AM-SU). Left: A composite spiking unit where astrocytes intertwine with neurons, influencing neural activity. Right: A schematic representation of astrocytes in tripartite synapses, modeling synaptic transmission and plasticity, and modulating neural activity. Below: The Astrocyte-Modulated Spiking Neural Network (AstroSNN) can empower large language models and give rise to conversational abilities.}
    \label{fig:bio}
\end{figure}

However, research within the computational modeling discipline has predominantly witnessed a paucity of explorations into neuron-astrocyte interactions. Most studies have been confined to modeling simple tasks~\cite{tsybina2022astrocytes, gordleeva2021modeling}  or theoretically comparing neuron-astrocyte communication with the architecture of artificial neural networks~\cite{kozachkov2023building}. These investigations have often not fully utilized the potential of astrocytes in complex pattern recognition tasks, nor have they delved deeply into their significant role in cognitive functions. Driven by the complex and synergistic interplay between neurons and astrocytes, we have led to the innovation and development of a groundbreaking computational model, termed the Astrocyte-Modulated Spiking Neural Network (AstroSNN). This model not only underscores the critical function of astrocytes in governing neural dynamics but also delineates the capacity of these cells to markedly enhance the efficiency of pattern recognition and cognitive processing. This enhancement is realized through the precise modulation of dynamic responses inherent within neural networks.

In this research, we introduce an innovative integration of astrocytes into SNNs, culminating in the development of the AstroSNN. This model intricately encapsulates the complex interactions inherent in neuronal activities, thereby enhancing the dynamic characteristics of neurons to more faithfully replicate biological neuronal behaviors. AstroSNN is distinguished not only by its substantial biological fidelity but also by its advanced computational capabilities. It effectively overcomes the longstanding challenge of implementing parallel training processes in traditional neuronal models, exhibiting versatility in transitioning between serial and parallel computational frameworks to meet diverse computational demands. Through rigorous empirical testing, AstroSNN has demonstrated exceptional proficiency in executing fundamental cognitive tasks crucial for the development of intelligence, particularly excelling in functions such as working memory and credit assignment, thereby surpassing the competencies of existing SNN models. Furthermore, upon the application of AstroSNN to complex, large-scale language generation tasks, empirical observations have indicated that the model exhibits performance on par with leading large-scale models. Notably, its enhanced efficacy becomes particularly pronounced in comparison with SNN-based architectures such as SpikeGPT~\cite{zhu2023spikegpt}. Comparative analysis with the Transformer architecture further indicates AstroSNN's reduced computational complexity and improved efficacy, reinforcing the benefits of incorporating astrocytic dynamics into contemporary neural network designs. This research significantly narrows the divide between biological neural systems and computational models, heralding a pivotal advancement in the realm of biologically inspired neural computing.

\section{Astrocyte-Modulated Spiking Unit}
Astrocytes, as an indispensable part of the central nervous system, form tripartite synapse structures with both pre- and post-synaptic neurons, actively engaging in inter-neuronal communication through the release of gliotransmitters. This regulatory mechanism plays a crucial role in the information processing capabilities of neural networks. In this study, we have integrated astrocytes into the SNNs. As illustrated in Figure~\ref{fig:bio}, neurons are interconnected via synapses, with the thickness of synapses representing the varying strengths of connections. In traditional SNN models, neurons form the foundational information processing network through these synapses. These synapses not only physically connect neurons but also play a vital role in transmitting and modulating neural signals, thus influencing the network's response and information processing capacity. Our study further incorporates astrocytes, which, through their connections with multiple neuronal synapses, add an additional, long-term layer of information processing to SNNs. The incorporation of astrocytes into SNNs markedly enhances the network's structural and functional complexity. These cells introduce additional dimensions to neural computation by modulating synaptic activities and neurotransmitter dynamics, thereby embedding a temporal aspect into the network's memory and information processing faculties. Crucially, the Astrocyte-Modulated Spiking Units (AM-SU) facilitate the SNNs' ability to form and sustain long-term associative memory between input signals, an attribute challenging to replicate in conventional neuron-centric network models. %This novel architectural integration offers substantial improvements in the network’s capacity for processing tasks involving long-term memory and intricate sequence analysis, representing a significant advancement in the field of neural network design.

% 我们在随后的章节中对 AM-LIF 进行精确的公式化描述，并阐明它与其他经典神经网络架构的关系。在 AM-LIF 中，相邻的神经元由星形胶质细胞紧密连接，因此我们在这里的重点是描述神经元-星形胶质细胞组合的集体动态特性，而不是像现有著作中通常所做的那样，只关注单个神经元的计算模型。为了清晰和简单起见，我们省略了当前层所有变量的层符号。神经元-胃泌素集合体对来自 $l-1$ 层的神经元活动的处理可表示为
% In the subsequent sections, we provide a rigorous formalization of the AM-LIF and elucidate its relationship with other classical neural network architectures.

\subsection{Computational modeling of AM-SU}

% 在 AM-SU 中，相邻的神经元通过星形胶质细胞紧密相连，这使得我们将重点放在神经元-星形胶质细胞集合的集体动态特性上，而不是像现有研究中常见的那样，只关注单个神经元的计算模型。为使表述清晰简洁，我们省略了所有当前层变量的层符号。由神经元-胃泌素形成的三方突触处理的来自 $l-1$ 层的神经元活动可以分为三个过程: 1. 突触前神经元的神经递质释放和对于突触后神经元的直接影响, 2. 神经递质对于星形胶质细胞的影响和星形胶质细胞对于不同突触的信息整合, 3. 星形胶质细胞对于突触后神经元的调控. 并可以表示如下：
In the AM-SU framework, we emphasize the intricate interconnectedness of neighboring neurons via astrocytes, thereby shifting the analytical focus towards the collective dynamic characteristics of the neuron-astrocyte ensemble. This approach diverges from the traditional focus on individual neuronal computational models predominant in extant literature. For the sake of clarity and brevity in our exposition, layer-specific symbols for all variables within the current stratum have been intentionally omitted. In this model, neuronal activity from the $l-1$ layer, as processed through the tripartite synapse by the neuron-astrocyte ensemble, encompasses a tri-phasic process:
\begin{itemize}
    \item The release of neurotransmitters by presynaptic neurons and their ensuing impact on postsynaptic neuronal activity;
    \item The astrocytic modulation of neurotransmitter dynamics and the consequential integration of synaptic information across disparate neuronal junctions;
    \item The astrocytic regulation of postsynaptic neuronal responses.
\end{itemize}

The details can be represented as follows:

\begin{align}
    \label{eq:current} & x_{t} = W_{x} s_{t}^{(l-1)},  \quad I_{t} = \sigma (x_{t}) % Neurotransmitter release and generation of post-synaptic current
\end{align}

% 方程~ref{eq:current}描述了神经网络中神经递质释放和突触电流产生的动态过程。第(l-1)层神经元在时间 $t$ 的尖峰输出由 $s_{t}^{(l-1)} \in \{ 0, 1 \}^N$表示，其中 $N$ 表示该层神经元的数量。以矩阵 $W_{x}$ 表示的突触权重决定了突触前神经元活动如何影响神经递质的释放，以 $x_t$ 表示。然后，神经递质浓度通过饱和函数 $\sigma(x_{t})$，将突触强度和神经递质浓度转换为输入电流 $I_{t}$。
Equation~\ref{eq:current} describes how neurotransmitters are released and create synaptic currents in the neural network. It shows spike outputs from the $(l-1)$th layer of neurons at time $t$, symbolized as $s_{t}^{(l-1)} \in \{ 0, 1 \}^N$ where $N$ is the neuron count. The synaptic weights, $W_{x}$, link presynaptic activity to neurotransmitter release, captured as $x_t$. These neurotransmitter levels then become input current $I_{t}$ through a function $\sigma(x_{t})$, reflecting synaptic strength and neurotransmitter concentration.

\begin{align}
    \label{eq:input}   k_{t} & = W_{k} x_{t}, \quad v_{t} = W_{v} x_{t}, \quad H_{t} = k_{t} v_{t}^T % Influence of neurotransmitters on astrocytes and integration of tripartite synapse information
\end{align}

% 方程~ref{eq:input}概括了神经递质如何影响星形胶质细胞，以及它们在三方突触间整合信号的作用。参数 $k_t$ 和 $v_t$ 分别量化了星形胶质细胞从当前和相邻突触吸收神经递质的情况。星形胶质细胞状态调制用 $\Delta H_t = k_{t} v_{t}^T$ 表示。这种建模方法将重点转移到了神经元-星形胶质细胞集合内的集体动力学上，正如隐藏状态矩阵 $H_t \in \mathbb{R}^{N \times N}$所捕获的那样，从而超越了对单个神经元的分析。该模型由此说明了星形胶质细胞介导的相邻神经元之间的相互影响机制。
Equation~\ref{eq:input} encapsulates how neurotransmitters affect astrocytes and their role in integrating signals across tripartite synapses. Parameters $k_t$ and $v_t$ respectively quantify the neurotransmitter absorption by astrocytes from current and adjacent synapses. The astrocytic state modulation is represented by $\Delta H_t = k_{t} v_{t}^T$. This modeling approach shifts the focus to the collective dynamics within the neuron-astrocyte ensemble, as captured in the hidden state matrix $H_t \in \mathbb{R}^{N \times N}$, thus transcending individual neuron analysis. The model thereby illustrates the mutual influence mechanism among neighboring neurons mediated by astrocytes.

\begin{align}
    \label{eq:output1} & q_{t} = W_{q} x_{t}, \quad o_{t} = H_{t} q_{t} % Influence of neurotransmitters on astrocyte read-out strength and regulation of post-synaptic neurons by astrocytes
\end{align}

% 方程~ref{eq:output1}描述了星形胶质细胞对突触读出强度（以 $q_{t}$ 表示）的影响，以及它们对突触后神经元膜电位的调节（以 $o_{t} = H_{t} q_{t}$ 表示）。三方突触内的星形胶质细胞相关过程被建模为线性函数，并与神经递质强度变量（$W_q$, $W_k$, $W_v in \mathbb{R}^{N \times N}$）相关联。这种线性建模方法能有效模拟星形胶质细胞对神经递质释放和突触整合的影响，同时兼顾计算效率和降低计算复杂度。
Equation~\ref{eq:output1} delineates the astrocytic influence on synaptic readout strength, denoted as $q_{t}$, and their regulation of post-synaptic neuron membrane potential, represented by $o_{t} = H_{t} q_{t}$. Astrocyte-related processes within tripartite synapses are modeled as linear functions, linked to neurotransmitter strength variables ($W_q$, $W_k$, $W_v \in \mathbb{R}^{N \times N}$). This linear modeling approach effectively simulates astrocytic impacts on neurotransmitter release and synaptic integration, balancing computational efficiency with a reduction in computational complexity.

% 引入三方突触和星形胶质细胞后，突触后状态的变化可表示为
% After the introduction of tripartite synapses and astrocytes, the postsynaptic state changes can be expressed as:
\begin{align}
    \label{eq:mem_update} u_{t} & = \frac{1}{\tau_n} u_{t-1} + I_{t}R                         \\
    \label{eq:attention} H_{t}  & = \frac{1}{\tau_a} H_{t-1} + \frac{k_{t} v_{t}^T}{\sqrt{N}} \\
    \label{eq:hidden} s_{t}     & = g(H_{t} q_{t} + u_{t})
\end{align}

% 在增强型 AM-SU 模型中，突触后神经元的动态是通过一组反映突触电流和星形胶质细胞调节影响之间相互作用的方程来捕捉的。神经元的膜电位 $u_{t}$随时间衰减，并受突触电流 $I_{t}$ 和膜电阻 $R$ 的调节。代表星形胶质细胞集体影响的隐藏状态矩阵 $H_{t}$ 根据等式~\ref{eq:attention}变化，其中 $\tau_a$ 表示星形胶质细胞时间常数，并捕捉受神经递质浓度影响的综合突触活动，这些神经递质浓度编码在 $k_{t}$ 和 $v_{t}$ 中。
In the AM-SU model, the dynamics of postsynaptic neurons are captured through a set of equations that reflect the interplay between synaptic currents and the modulating influence of astrocytes. The membrane potential of a neuron $u_{t}$, which decays over time and is modulated by synaptic current $I_{t}$ and membrane resistance $R$, is updated as shown in Equation~\ref{eq:mem_update}. The hidden state matrix $H_{t}$, representing the collective astrocytic influence, evolves according to Equation~\ref{eq:attention}, where $\tau_a$ denotes the astrocytic time constant and captures the integrated synaptic activity influenced by neurotransmitter concentrations encoded in $k_{t}$ and $v_{t}$.

% 突触和星形胶质细胞活动的整合为尖峰状态 $s_{t}$ 的更新提供了信息，如方程~ref{eq:hidden}所述。我们使用简单的相加的形式将星形胶质细胞状态$H_{t}$与膜电位$u_{t}$合并，从而产生一种新的尖峰状态，这种状态既体现了直接的突触输入，也体现了星形胶质细胞介导的突触整合所提供的更广泛的背景。这一框架使 AM-SU 模型能够支持复杂的时序处理，保持扩展的记忆轨迹，有选择地过滤和保留与当前计算任务相关的信息。
This integration of synaptic and astrocytic activity informs the update of the spiking state $s_{t}$ as outlined in Equation~\ref{eq:hidden}. The function $f(\cdot, \cdot)$ merges the astrocytic state $H_{t}$ with the membrane potential $u_{t}$, leading to a new spiking state that embodies both the immediate synaptic input and the broader context provided by astrocyte-mediated synaptic integration. This framework allows the AM-SU model to support complex temporal processing by maintaining an extended memory trace and enabling the selective filtering and retention of information pertinent to the computational task at hand.

% 函数$g(\cdot)$表示AM-SU内部状态到脉冲活动的映射. 在这项工作中，神经元的输出函数被建模为将集合的隐藏状态添加到膜电位的简单二值化。这种二值化反映了生物学上观察到的动作电位阈值机制，即神经元在超过特定电压阈值 $v_{th}$ 时会发射。所提出的框架捕捉了神经元和星形胶质细胞之间的双向信息流和动态交互，从而丰富和增强了网络的表征和适应性。星形胶质细胞的加入提供了一个更精确的生物模型，并可能赋予神经网络架构以前未被发现的计算优势。
The function $g(\cdot)$ represents the mapping from the internal state of AM-SU to pulse activities. In this work, the neuron's output function is modeled as a simple binarization of the ensemble's hidden state added to the membrane potential, as shown in Equation~\ref{eq:output}. This binarization reflects the biologically observed action potential threshold mechanism, whereby a neuron will fire when exceeding a specific voltage threshold $v_{th}$. The proposed framework captures the bidirectional information flow and dynamic interactions between neurons and astrocytes, thereby enriching and enhancing the network's representations and adaptability. Incorporating astrocytes offers a more accurate biological model and might confer computational advantages previously undiscovered in neural network architectures.

\begin{equation}
    \begin{aligned}
        g(x) & = \begin{cases}
                     1, \quad x \geq v_{th} \\
                     0, \quad x < v_{th}
                 \end{cases}
    \end{aligned}
    \label{eq:output}
\end{equation}

\subsection{AM-SU: Towards Efficient Neuronal Modeling}
\label{sec:as_rnn}

% 脉冲神经网络常被认为是循环神经网络的子集, 根据外部输入以及上一时刻的内部状态更新输出以及当前的内部状态. 对比LIF / ALIF / LSTM / AM-LIF  -> 1. 神经元标量, 无法communicate. 2. LSTM 没办法parallel  3. AM-LIF 具有更多的hidden state表示.

% 递归神经网络（RNN）一直是我们理解时序处理和序列数据建模不可或缺的一部分。尖峰神经网络（SNN）通常被视为RNN的一个子集，因为它们具有根据外部输入和先前状态更新输出和当前内部状态的固有能力。在本小节中，深入比较分析几种神经元模型，如 LIF、ALIF和拟议的 AM-LIF。

% Recurrent Neural Networks (RNNs) have always been fundamental to our comprehension of time series processing and sequence data modeling. Spiking Neural Networks (SNNs), due to their inherent ability to update their outputs and current internal states based on external stimuli and prior states, are often perceived as a subset of RNNs. In this subsection, we will examine various neuron models, including LIF, ALIF~\cite{yin2021accurate}, and the proposed AM-SU.

\begin{figure}[htbp]
    \centering
    \includegraphics[width=\linewidth]{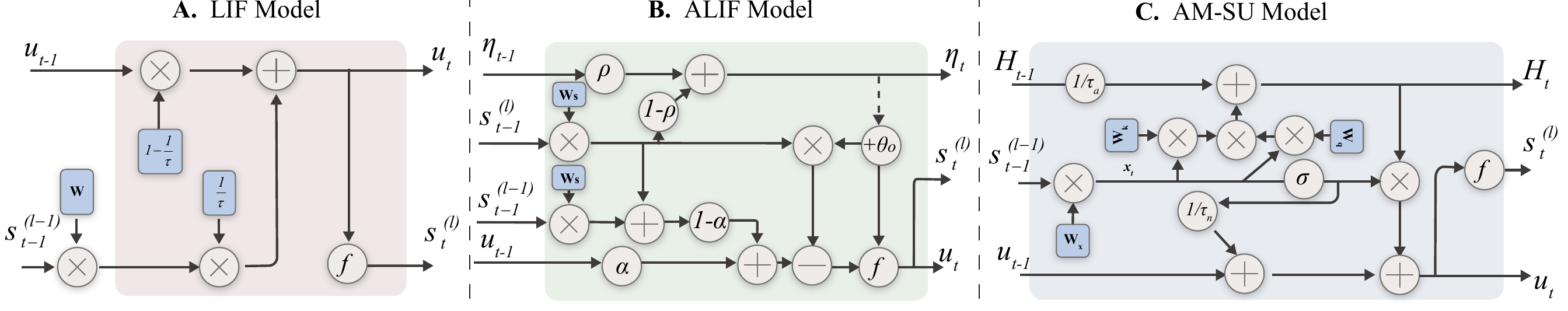}
    % 计算模型 LIF、ALIF 和 AM-SU 的图示。A. LIF 模型展示了尖峰神经元的基本机制，即膜电位\( u_{t-1} \)通过突触权重\( W \)根据前一状态和输入尖峰\( S_{t-1} \)进行更新。B. ALIF 模型通过加入自适应阈值（theta）和衰减因子（rho）来增强 LIF，阈值会根据神经元的发射历史（如 \( \eta_{t-1} \)和衰减因子 \( \rho \)所示）进行调整。此外，AM-SU 模型还引入了自适应连接，使得当前层的前一个尖峰（s^{(l)}_{t-1}）也可以作为神经元当前状态的输入。C. AM-SU 模型结合了星形胶质细胞调制，星形胶质细胞的状态\( H_{t-1} \)和神经元的膜电位\( u_{t-1} \)以协调的方式更新，根据星形胶质细胞和神经元的综合状态影响尖峰输出\( S_{t}^{(l)} \)。与前几种模型不同的是，AM-SU 的内部状态只有线性状态转换，因此具有并行化的潜力。
    \caption{Illustration of the computational models LIF, ALIF, and AM-SU. A. The LIF model demonstrates the fundamental mechanism of spiking neurons where the membrane potential \( u_{t-1} \) is updated based on the preceding state and the input spike \( S_{t-1} \) through synaptic weights \( W \). B. The ALIF model enhances the LIF by incorporating an adaptive threshold \( \theta \), which adjusts according to the neuron's firing history. Additionally, it introduces autaptic connections, enabling the previous spike \( s^{(l)}_{t-1} \) from the current layer to also serve as an input to the neuron's current state. C. The AM-SU model incorporates astrocytic modulation where the astrocyte state \( H_{t-1} \) and the neuron's membrane potential \( u_{t-1} \) are updated in a coordinated fashion, influencing the spiking output \( S_{t}^{(l)} \) based on the combined state of astrocytes and neurons. Unlike the former models, the AM-SU features only linear state transitions within its internal states, offering potential for parallelization.}
    \label{fig:comp}
\end{figure}

% 如图~\ref{fig:comp}所示，我们将这些计算模型进行了比较。传统的神经元模型（如 LIF 和 ALIF）虽然受到生物启发，但对真实世界的神经元行为进行了相当简化的抽象。它们的主要局限是将神经元表示为标量实体, 具体体现在神经元的隐状态不存在向量操作. 这种模型导致神经元无法对同一层的神经元的活动产生影响。这种方法阻碍了神经元之间的交流，因为它们无法有效传递一系列数值或复杂信息, 这与生物神经系统大相径庭。这种标量性质虽然计算效率高，但却牺牲了丰富的表征和适应性，从而限制了网络捕捉复杂模式和依赖关系的能力。相比之下, ALIF则通过将神经元的脉冲也视为一种内部状态, 来缓解神经元的通信限制挑战. ALIF为同一层的神经元分配了自突触, 使神经元能够将脉冲活动广播到同一层的所有神经元, 实现了同一层神经元的交流.
As depicted in Figure~\ref{fig:comp}, conventional neuron models like the Leaky Integrate-and-Fire (LIF) model, despite their biological inspiration, present a simplified abstraction of the multifaceted behaviors observed in actual neurons. These models primarily conceptualize neurons as scalar entities, a limitation prominently reflected in their inability to perform vector operations within neuronal hidden states. This representation impedes the capacity of neurons to exert influence on their contemporaries within the same layer, thereby hindering the transmission of diverse values or complex information—a stark deviation from the rich communicative dynamics of biological neural systems. While such scalar-based approaches yield computational efficiency, they impose significant limitations on the network's ability to capture and process complex patterns and dependencies. In contrast, the Adaptive LIF (ALIF) model~\cite{yin2021accurate} addresses these communicative constraints by conceptualizing neuron spikes as an integral component of their internal state. ALIF enhances intralayer neuronal communication by assigning autapses to neurons, enabling the dissemination of spiking activities across the entire layer, thereby facilitating a more effective and biologically plausible mode of neuronal interaction. While our AM-SU leverages the integration of astrocytes to enhance its computational framework. This integration results in an expanded hidden state representation, $H_t$, scaled quadratically with the number of neurons, thereby markedly augmenting the model's memory and data processing faculties. This innovative approach, which synergistically combines the dynamic temporal attributes of spiking neurons with the regulatory effects of astrocytes, enables AM-SU to encompass a wider array of neuronal activities and dependencies.

The ability to support temporal unfolding and parallelized training is crucial for advancing towards large-scale language models~\cite{vaswani2017attention}. Figure\ref{fig:comp} b\&c illustrates how LIF and ALIF neuron models, with their inherent complex nonlinear operations, enable selective memory storage, forgetting, and regulation of neuron output. However, this complexity results in a dependency of the neuron's hidden state on preceding temporal states, impeding the possibility of time-independent model unfolding for parallelized training. Such constraints considerably limit the efficiency of training SNNs and pose challenges to the development of large-scale SNNs. In contrast, the AM-SU model introduces linear state transitions in both $u_t$ and $H_t$~\cite{orvieto2023resurrecting}. This design, which integrates the neuron's membrane potential with astrocytes' hidden states, circumvents complex nonlinear state alterations while preserving the model's comprehensive representational capacity and functional efficacy. Crucially, AM-SU's linear transformations enable efficient parallelized computations within SNNs, marking a significant advancement in neural network training methodologies.
% Supporting temporal unfolding and parallelized training is a prerequisite for moving towards large language models~\cite{vaswani2017attention}. As illustrated in Figure~\ref{fig:comp} b\&c, the complex nonlinear operations within LIF and ALIF neuron models enable selective memory storage, forgetting, and regulation of neuron output. However, this also results in the neuron's hidden state being strictly dependent on previous moments. This dependency prevents the neuron model from being unfolded independently over time for parallelized training, significantly limiting the training efficiency of SNNs and hindering the development of large-scale SNNs. In contrast, the state transitions of hidden states $u_t$ and $H_t$ in AM-SU are linear operations~\cite{orvieto2023resurrecting}. This design, which considers both the neuron's membrane potential and the hidden state of astrocytes, avoids nonlinear changes in the collective hidden state of the cells, while ensuring the model's rich representation and performance. More importantly, such linear transformations of hidden states in AM-SU enable parallelized computations in SNNs.

% 在AM-SU中，不同时刻的内部状态线性相关, 非线性的操作仅存在于对于输入和输出的处理. 这样的设计实现了AM-LIF内部状态的时间上的解耦, 以星形胶质细胞的隐状态$H_{t}$为例:
In AM-SU, states internal to different moments are linearly correlated, with nonlinear operations confined to handling inputs and outputs. Such a design effectuates a temporal decoupling of AM-SU's internal states, exemplified by the astrocytic hidden state $H_{t}$:

\begin{equation}
    \begin{aligned}
        H_{t} & = \frac{1}{\tau_a} H_{t-1} + \frac{k_{t} v_{t}^T}{\sqrt{N}}                                                                                                                                                  \\
              & = \frac{1}{\tau_a^2} H_{t-1} + \frac{1}{\tau_a} \frac{k_{t-1} v_{t-1}^T}{\sqrt{N}} + \frac{k_{t} v_{t}^T}{\sqrt{N}}                                                                                          \\
              & = \frac{1}{\tau_a^t} \frac{k_{0} v_{0}^T}{\sqrt{N}} + \frac{1}{\tau_a^{t - 1}} \frac{k_{1} v_{1}^T}{\sqrt{N}} + \dots + \frac{1}{\tau_a} \frac{k_{t-1} v_{t-1}^T}{\sqrt{N}} + \frac{k_{t} v_{t}^T}{\sqrt{N}} \\
              & = \sum_{t^\prime=0}^{t^\prime=t} \frac{1}{\tau_a^{t-t^\prime}} \frac{k_{t^\prime} v_{t^\prime}^T}{\sqrt{N}}
    \end{aligned}
    \label{eq:h_sum}
\end{equation}

% 如式~\ref{eq:eq:h_sum}, 在知道之前时刻所有的输入刺激, 就能够通过O(1)的复杂度并行化计算出之前所有时刻的神经元的输出, 而小节~\ref{sec:as_rnn}中讨论的其他神经元模型则都因为隐状态的转移经过了非线性操作, 导致隐状态无法被直接表示为输入刺激的求和. 将神经元的隐状态表示为如式~\ref{eq:h_sum}的好处就是可以将不同神经元空间上的注意力并行化计算, 转变成时间上的注意力:
As exemplified by Equation $\ref{eq:h_sum}$, given inputs from prior moments, neuronal outputs across all those moments can be computed in parallel, with complexity $O(1)$. This stands in contrast to other neuronal models shown in Figure~\ref{fig:comp}, where nonlinear transitions of hidden states prevent the direct representation of hidden states as summations over input stimuli. Expressing neuronal hidden states as in Equation $\ref{eq:h_sum}$ bears the advantage of parallelizing attention across different neuronal spaces over time:

\begin{equation}
    \begin{aligned}
        X        = S^{(l-1)} W_x, & \quad K  = X W_k, \quad V = X W_v, \quad Q = X W_q               \\
        S                         & = g(\frac{(Q K^T \odot M^{(a)})V}{\sqrt{N}} + RM^{(n)}\sigma(X)) \\
        M^{(p); i,j}              & = \begin{cases}
                                          \tau_{p}^{j-i}, \quad i \geq j \\
                                          0, \quad  i < j                \\
                                      \end{cases}, \quad p \in \{a, n\}
    \end{aligned}
    \label{eq:parallel}
\end{equation}

% 如式~\ref{eq:parallel}所示, 将AM-LIF的时间常数扩展为不同时刻对应关系的矩阵形式, 就能够实现具有AM-LIF的SNN的并行化计算. 其中$(p)$代表了不同类别的细胞, 包括神经元和星形胶质细胞. 总之，虽然 Transformer 在许多任务中（尤其是 NLP）都树立了标杆，但 AM-LIF 架构可以被认为是一种特殊的Transformer结构,受到星形胶质细胞的启发, 将注意力机制融入到了神经元模型之中, 将空间注意力与递归特性融为一体，在计算效率和并行化能力方面具有优势，有可能成为特定应用的合适选择。
In Equation~\ref{eq:parallel}, the temporal constant in the AM-LIF model is expanded into a matrix $M^{(p)}$, facilitating parallel computation in SNNs incorporating this model.  $(p)$ denotes various cell types, including neurons $(n)$ and astrocytes $(a)$. Concluding, while the Transformer architecture has established itself as a benchmark in diverse tasks, the AM-SU can be considered a specialized variant, drawing inspiration from astrocytic functions. It incorporates an attention mechanism directly within the neuronal model, harmonizing spatial attention with recursive dynamics. This integration offers distinct advantages in computational efficiency and parallelization, positioning the AM-LIF as a potentially optimal architecture for specific computational tasks.

\section{Results}

% 在本节中，我们对 AstroSNN 在一系列任务中的表现进行了严格评估，这些任务从简单到复杂，包括典型的神经科学基准任务，如工作记忆任务和学分分配任务，以及自然语言生成。我们的研究结果表明，AM-SU 能有效增强网络的记忆和联想能力。在此基础上，我们以 AstroSNN 为骨干，大胆地训练了一个大型语言模型，发现了它的新兴会话能力，并在零拍任务中取得了出色的表现。此外，我们还对使用 AstroSNN 建立的大型语言模型的推理开销和可视化进行了分析，从而证实了其效率。本节将详细介绍这些实验，说明 AstroSNN 在各种复杂任务中的强大能力和广泛应用。
In this section, we rigorously evaluate the performance of AstroSNN across a range of tasks, from the simple to the complex, encompassing typical neuroscience benchmarks such as working memory tasks, as well as natural language generation. Our findings demonstrate that AM-SU effectively enhances the network's memory and associative capabilities. Building on this, we ventured to train a large language model using AstroSNN as the backbone, uncovering its emergent conversational abilities and achieving remarkable performance on zero-shot tasks. Further, we conducted an analysis of the inference overhead and visualization for the large language model built with AstroSNN, substantiating its efficiency. %This section details these experiments, illustrating the robust capabilities and versatile applications of AstroSNN in various complex tasks.

\subsection{Working Memory Experiments}

% 在实验中，我们使用两个不同的任务评估了不同神经元模型的性能，包括LIF、广义LIF（ALIF）和AM-SU，这两个任务分别是记忆长度任务和伞状长度任务。记忆长度任务是一个具有单一二进制情境的T迷宫，所需的记忆长度从1对数增加到1500。而 "伞状长度 "任务则是一个 "伞状问题"，即在第一个决定的重要性之后会出现一连串的干扰变量，而这些变量的长度也是按对数从1增加到1500。
Here, we evaluated the performance of different neuronal models, including LIF, Adaptive LIF (ALIF)~\cite{yin2021accurate}, and AM-SU, using two distinct tasks: the Memory Length task and the Umbrella Length task. The Memory Length task involved a T-maze~\cite{o1971hippocampus} with a single binary context, where the required memory length was logarithmically increased from 1 to 1500. The Umbrella Length task presented a 'umbrella problem' where the importance of the first decision was followed by a chain of confounding variables, with the chain length also varying logarithmically from 1 to 1500.

\begin{figure}[htbp]
    \centering
    \includegraphics[width=\linewidth]{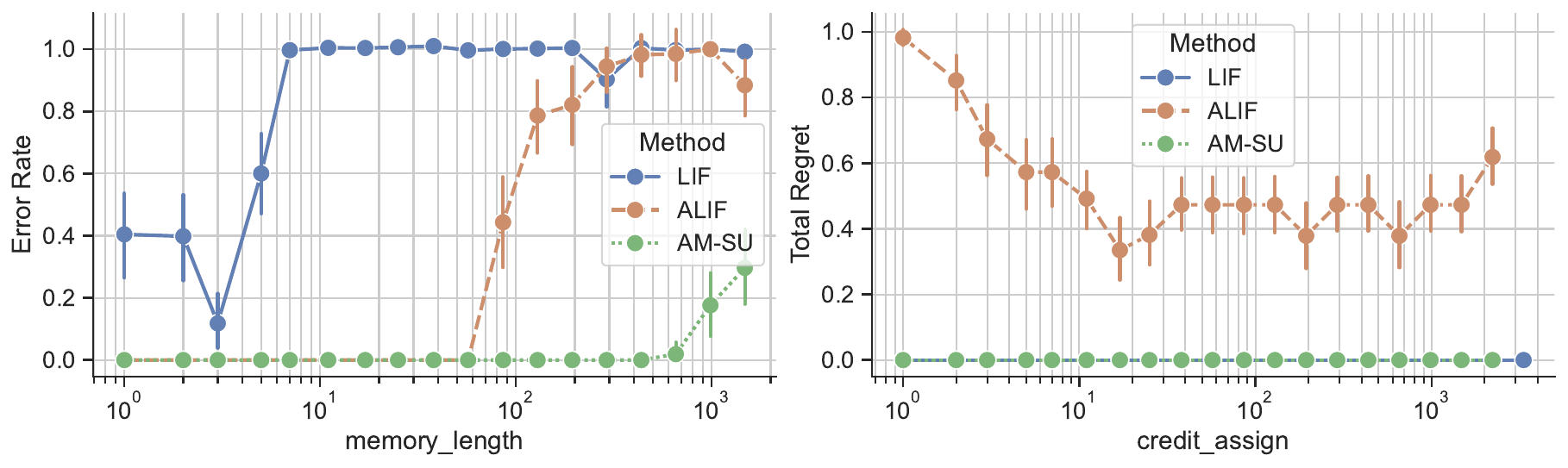}
    % 神经元模型在不同记忆和决策情境中的表现比较。左图：在 T 型迷宫任务中，LIF、ALIF 和 AM-SU 模型的错误率与记忆长度的函数关系，记忆长度从 1 到 1500 对数缩放。右图 在长度从 1 到 1500 对数变化的混杂变量任务中，每个模型在不同伞长度下的总后悔率。AM-SU 模型的错误率和后悔率都明显较低，表明其记忆保持和决策效率更高
    \caption{Comparative performance of neuronal models across different memory and decision-making contexts. Left: Error rates of LIF, ALIF, and AM-SU models as a function of memory length in a T-maze task. Right: Total regret measures for each model across umbrella lengths in a confounding variable task.}%, with lengths varying logarithmically from $1$ to $1500$.} %The AM-SU model demonstrates markedly lower error rates and regret, indicating superior memory retention and decision-making efficacy.}
    \label{fig:mem_len}
\end{figure}

% 结果如图~ref{fig:mem_len}所示，显示了每个模型的不同性能特征。在 "内存长度 "任务中，LIF 模型在所有长度上都表现出较高的错误率，而 ALIF 模型则在错误率显著增加之前提高到一个临界值。相比之下，AM-SU 模型始终保持较低的错误率，显示出其在较长时间内强大的记忆保持能力。同样，在 "伞状长度 "任务中，LIF 模型在处理混淆信息时很吃力，总遗憾率很高。ALIF 模型的情况较好，随着序列变长，总遗憾率有所下降。然而，AM-SU 模型表现突出，遗憾最小，它有效地将注意力集中在关键的初始决策上，并有效地过滤掉了无关的后续信息。
The results, as shown in Figure~\ref{fig:mem_len}, reveal distinct performance characteristics of each model. In the Memory Length task, while the LIF model exhibited high error rates across all lengths, the ALIF model improved up to a threshold before its error rate increased significantly. In contrast, the AM-SU model consistently maintained low error rates, showcasing its robust memory retention capabilities over extended periods. Similarly, in the Umbrella Length task, the LIF model struggled with confounding information, displaying high total regret. The ALIF model fared better, showing a decrease in total regret with longer sequences. However, the AM-SU model stood out with minimal regret, efficiently maintaining focus on the crucial initial decision and effectively filtering out irrelevant subsequent information.

% 实验结果有力地验证了 AM-SU 模型在处理长期记忆和决策干扰因素方面的卓越性能。受星形胶质细胞启发而设计的 AM-SU 在处理长期依赖关系和关注关键时间事件方面超越了 LIF 和 ALIF 等传统模型。这些发现强调了将类似星形胶质细胞的机制纳入神经元模型的显著优势，增强了它们处理复杂时间的能力，使它们在高级神经计算任务中非常有效。
The experimental outcomes robustly validate the AM-SU model's superior performance in handling long-term memory and decision-making amid confounding factors. Its astrocyte-inspired design allows AM-SU to surpass traditional models like LIF and ALIF in processing long-term dependencies and focusing on key temporal events. These findings underscore the significant benefits of incorporating astrocyte-like mechanisms into neuronal models, enhancing their ability for complex temporal processing and rendering them highly effective for advanced neural computing tasks.

% 与其他的神经元模型相比, AM-SU能够有效地提高网络的记忆和联想能力. 通过引入星形胶质细胞, AM-SU模型在处理长期依赖关系和关注关键时间事件方面超越了传统的LIF和ALIF模型. 这些发现强调了将类似星形胶质细胞的机制纳入神经元模型的显著优势. 因此, 接下来我们将尝试将AM-SU模型应用更大的模型中, 并构建星形细胞调制的脉冲神经网络, 并在自然语言处理任务中进行评估.
Compared to conventional neuron models, the AM-SU demonstrates an enhanced ability to augment network memory and associative capabilities. These findings underscore the significant advantages of integrating astrocyte-like mechanisms into neuronal models. Thus, moving forward, we aim to apply the AM-SU model to larger frameworks and construct AstroSNN for evaluation in natural language processing tasks.%Incorporating astrocytic mechanisms enables the AM-SU model to surpass traditional LIF and ALIF models in processing long-term dependencies and focusing on key temporal events. 

\subsection{Natural Language Generation}

% 评估语言模型生成连贯且与上下文相关文本的能力是自然语言生成（NLG）的一个重要方面。为此，我们将利用尖峰神经元的 AstroSNN 模型与一系列传统和先进的神经网络模型进行了比较。这种比较基于它们的性能指标，如在 enwiki8 数据集上的每字符比特数（BPC），用于衡量字符级预测效率；在 WikiText-103 数据集上的每混淆度（PPL），用于评估单词级预测能力。
Evaluating the proficiency of language models in generating coherent and contextually relevant text is an essential aspect of Natural Language Generation (NLG). To this end, we compared our AstroSNN models with a range of traditional and advanced neural network models. This comparison was based on their performance metrics, such as bits-per-character (BPC) on the enwiki8 dataset for character-level prediction efficiency, and perplexity (PPL) on the WikiText-103 dataset for assessing word-level prediction capability.
\begin{table}[ht]
    \centering
    \caption{Natural Language Generation Performance Metrics}
    \begin{tabular}{lcccccc}
        \hline
        \textbf{Method}                          & \textbf{Spiking} & \textbf{\tabincell{c}{Params                             \\ (M)}} & \textbf{L} & \textbf{d} & \textbf{\tabincell{c}{enwiki8 \\ (BPC)}} & \textbf{\tabincell{c}{WikiText-103 \\ (PPL)}} \\
        \hline
        Transformer~\cite{vaswani2017attention}  & \ding{55}        & 148M                         & 16 & 410  & -     & 26.1  \\
        Transformer~\cite{vaswani2017attention}  & \ding{55}        & 38M                          & 12 & 512  & 1.14  & -     \\
        Stacked LSTM~\cite{graves2013generating} & \ding{55}        & -                            & 7  & -    & 1.67  & -     \\
        \hline
        GPT-2 Small~\cite{radford2019language}   & \ding{55}        & 117M                         & 12 & 768  & 1.16  & 37.5  \\
        GPT-2 Medium~\cite{radford2019language}  & \ding{55}        & 345M                         & 24 & 1024 & 1.01  & 26.37 \\
        GPT-2 Large~\cite{radford2019language}   & \ding{55}        & 762M                         & 36 & 1280 & 0.97  & 22.05 \\
        \hline
        SpikeGPT~\cite{zhu2023spikegpt}          & \checkmark       & 45M                          & 12 & 512  & 1.283 & -     \\
        SpikeGPT~\cite{zhu2023spikegpt}          & \checkmark       & 216M                         & 18 & 768  & -     & 39.75 \\
        \hline
        AstroSNN Small (Ours)                    & \checkmark       & 88M                          & 12 & 512  & 1.142 & 32.97 \\
        AstroSNN Base (Ours)                     & \checkmark       & 306M                         & 16 & 1024 & 0.994 & 27.56 \\
        AstroSNN Large (Ours)                    & \checkmark       & 732M                         & 20 & 1536 & 0.987 & 21.83 \\
        \hline
    \end{tabular}
    \label{tab:nlg_performance}
\end{table}

% 表{/ref{tab:comprehensive_nlg_performance}中的结果表明，与传统神经网络（包括 Transformer 和 GPT-2 模型的各种配置）相比，我们的 AstroSNN 模型表现出更低的 BPC 分数和具有竞争力的 PPL 值。值得注意的是，AstroSNN 模型只需较少的参数就能获得这些结果，这突出了尖峰神经网络在语言建模任务中的效率。
The results in Table~\ref{tab:nlg_performance} demonstrate that our AstroSNN models exhibit lower BPC scores and competitive PPL values when juxtaposed with traditional neural networks, including various configurations of the Transformer and GPT-2 models. Notably, the AstroSNN models achieve these results with fewer parameters, nderscoring the efficiency of spiking neural networks augmented with astrocytes in language modeling tasks.

% 此外，比较结果表明，我们的模型在与同样使用尖峰神经元的 SpikeGPT 模型的比较中占据了优势。AstroSNN 模型在 BPC 中的表现优于 SpikeGPT，在 PPL 中也显示出良好的效果，这表明我们将尖峰神经元与语言建模技术相结合的方法非常有效。
Furthermore, the comparison illustrates that our models hold their ground against SpikeGPT models, which also utilize spiking neurons. The AstroSNN architectures exhibit superior performance over SpikeGPT in terms of Bits Per Character (BPC), and yield encouraging outcomes in Perplexity (PPL), indicating that our methodology of amalgamating spiking neurons enhanced with astrocytic mechanisms with advanced language modeling strategies is profoundly efficacious.

% AstroSNN 模型在 NLG 方面取得的进步为今后研究尖峰神经网络的可扩展性铺平了道路。随着计算成本的降低和处理效率的提高，尖峰神经网络（如 AstroSNN 模型中所采用的尖峰神经网络）可能会彻底改变 NLG 的面貌，尤其是在资源有限、模型效率至关重要的环境中。
The advancements in NLG as demonstrated by the AstroSNN models pave the way for future research into the scalability of SNNs. With the potential for reduced computational costs and efficient processing, SNNs like those employed in AstroSNN models could revolutionize the landscape of NLG, especially in resource-constrained environments where model efficiency is paramount.

% 基于这些见解，我们推测，扩大模型的规模并在大量数据集上进行预训练有可能激发会话能力。为了验证这一假设，我们在以多样性和复杂性著称的综合语料库 Pile 数据集上训练了一个拥有 15 亿个参数的 AstroSNN 模型。随后，我们在专为对话系统定制的 MOSS finetune 数据集上进行了微调。如下图所示，我们展示了一些对话样本。更多示例请参见章节~ref{sec:examples}。同时我们提供了一个在线对话的demo, 如脚注所示。
Building on these insights, we postulate that expanding the model's scale and pre-training on a substantial dataset could potentially elicit conversational abilities. To test this hypothesis, we trained an AstroSNN model with 1.5 billion parameters on the Pile dataset~\cite{pile}, a comprehensive corpus known for its diversity and complexity. Subsequent fine-tuning was conducted on the MOSS finetune dataset~\cite{sun2023moss}, which is tailored for dialogue systems.  As depicted below, we present some sample dialogues. For more examples, please refer to Supplementary Material. Additionally, we provide an online dialogue demo~\footnote{\url{http://mind-engine.org/mind-talk}}.

\begin{dialoguebox}[Philosophical Questions \#1]
    \textbf{Human:} What do you think is the meaning of life?\\
    \textbf{Mind Talk:}  As an AI language model, I don't have personal opinions or beliefs. However, I can provide you with some information on the meaning of life. The meaning of life is a philosophical question that has been debated by scientists, philosophers, and religious leaders for centuries. Some believe that the meaning of life is to find happiness and fulfillment in life, while others believe that it is to find meaning and purpose in life. Ultimately, the meaning of life is a complex and multifaceted concept that is open to interpretation $\dots...$
\end{dialoguebox}

% \begin{parcolumns}[colwidths={1=0.4\textwidth}, distance=1.5em]{2}
%     \colchunk[1]{\textbf{Human:}}
%     \colchunk[2]{How's the weather today?}  
%     \colplacechunks

%     \colchunk[1]{\textbf{Mind Talk:}}
%     \colchunk[2]{It's quite sunny with a gentle breeze, a perfect day for a walk outside.}
%     \colplacechunks

%     % ... Continue with the rest of the dialogue
% \end{parcolumns}

% \subsection{Zero-Shot Evaluation}  

% 基于 Kocon 等人的见解~\cite{kocon2023chatgpt}，我们探索了在 The Pile 数据集~\cite{pile}上训练的具有 15 亿个参数的放大 AstroSNN 模型的零点学习能力。零点学习任务测试了模型在没有特定训练的情况下对未知任务的泛化能力，这对语言模型的实际应用至关重要。
Building on Kocon et al.'s insights~\cite{kocon2023chatgpt}, we explored the zero-shot learning capabilities of the scaled-up AstroSNN model with 1.5 billion parameters, trained on The Pile dataset~\cite{pile}. Zero-shot learning tasks, which test the model's generalization to unseen tasks without specific training, are crucial for language models' practical application.

\begin{table}[ht]
    \centering
    \caption{Zero-Shot Learning Task Performance Comparison}
    \begin{tabular}{lc|ccccccc}
        \hline
        \textbf{Method} & \textbf{Spiking} & \textbf{PIQA} & \textbf{SciQ} & \textbf{WG} & \textbf{COPA} & \textbf{BoolQ} & \textbf{HS} & \textbf{Logiqa} \\
        \hline
        BLOOM           & \ding{55}        & 64.2          & 80.7          & 52.0        & 62.0          & 53.3           & 13.6        & 22.6            \\
        BERT            & \ding{55}        & 51.5          & 20.5          & 50.1        & 58.0          & 38.7           & 24.7        & 21.0            \\
        GPT2            & \ding{55}        & 69.2          & 80.4          & 55.3        & 72.0          & 60.5           & 36.4        & 22.6            \\
        Pythia          & \ding{55}        & 62.8          & 75.2          & 51.1        & 69.0          & 57.9           & 28.7        & 19.8            \\
        \hline
        SpikeGPT        & \checkmark       & 54.3          & 21.7          & 51.8        & 51.0          & 38.4           & 25.8        & 20.1            \\
        AstroSNN 1B5    & \checkmark       & 68.4          & 78.7          & 53.4        & 69.7          & 57.8           & 36.7        & 22.7            \\
        \hline
    \end{tabular}
    \label{tab:zero_shot_performance}
\end{table}

% 我们评估了 AstroSNN 在各种数据集上的性能，包括 Piqa~\cite{bisk2020piqa}、Sciq~\cite{welbl2017crowdsourcing}、WinoGrande~\cite{sakaguchi2021winogrande}、Copa~\cite{clark2019boolq}、 BoolQ~\cite{clark2019boolq}, HellaSwag~\cite{zellers2019hellaswag}, and LogiQA~\cite{liu2020logiqa}, benchmarking it against large-scale non-spiking and spiking models (Table ~/ref{tab:zero_shot_performance}）。结果证实了 AstroSNN 强大的零点学习能力，其性能在与大型非尖峰模型的比较中极具竞争力，并在大多数任务中超过了 SpikeGPT。这凸显了像 AstroSNN 这样的尖峰模型在规模上的潜力，有可能在零点学习方面超越传统的深度学习模型。
We assessed AstroSNN's performance across various datasets, including Piqa~\cite{bisk2020piqa}, Sciq~\cite{welbl2017crowdsourcing}, WinoGrande~\cite{sakaguchi2021winogrande}, Copa~\cite{clark2019boolq}, BoolQ~\cite{clark2019boolq}, HellaSwag~\cite{zellers2019hellaswag}, and LogiQA~\cite{liu2020logiqa}, benchmarking it against large-scale non-spiking and spiking models (Table~\ref{tab:zero_shot_performance}). The results confirm AstroSNN's robust zero-shot learning capability, showcasing competitive performance against large non-spiking models and surpassing SpikeGPT in most tasks. This underscores the potential of spiking models like AstroSNN at scale, potentially exceeding traditional deep learning models in zero-shot learning.

\subsection{Efficiency Analysis}

% 自然语言生成系统的效率，尤其是在推理过程中的效率，受到其并行化训练能力以及时间和空间复杂性的严重影响。本节将对包括 AstroSNN 在内的各种架构进行比较分析，重点介绍它们在复杂性和性能指标（如延迟、吞吐量和内存使用量）之间的平衡。

%The efficiency of natural language generation systems, particularly during inference, is critically influenced by their ability to parallelize training, as well as their time and space complexities. 

In this section, we present a comparative analysis of diverse architectures, including AstroSNN, with an emphasis on evaluating their trade-offs between computational complexity and key performance metrics like latency, throughput, and memory utilization.

\begin{table}[htbp]
    \centering
    % 不同架构在并行化、时间复杂性、空间复杂性和性能方面的比较。
    \caption{Comparison of different architectures in terms of parallelization, time complexity, space complexity, and performance.}
    \label{tab:architecture_comparison}
    \begin{tabular}{lcccc}
        \hline
        \textbf{Architectures} & \textbf{Parallelization} & \textbf{\tabincell{c}{Time                                               \\ Complexity}} & \textbf{\tabincell{c}{Space \\ Complexity}} & \textbf{Performance} \\
        \hline
        Transformer            & \checkmark               & \(O(T^2d)\)                & \(O(T^2 + Td)\)      & \checkmark\checkmark \\
        Reformer               & \checkmark               & \(O(T \log T \cdot d)\)    & \(O(T \log T + Td)\) & \checkmark           \\
        Linear Transformers    & \checkmark               & \(O(Td^2)\)                & \(O(Td + d^2)\)      & \checkmark           \\
        SpikeGPT               & \checkmark               & \(O(Td)\)                  & \(O(d)\)             & \checkmark           \\
        Recurrent NN           & \ding{55}                & \(O(Td)\)                  & \(O(d)\)             & \ding{55}            \\
        AstroSNN               & \checkmark               & \(O(Td)\)                  & \(O(d^2)\)           & \checkmark\checkmark \\
        \hline
    \end{tabular}
\end{table}

% 并行化对于扩展训练过程至关重要，Transformer、Reformer、Linear Transformers、SpikeGPT 和 AstroSNN 等架构都支持并行化，从而大大缩短了训练时间。其中，Reformer 擅长处理较长的序列，因为它的时间复杂度为（O(T \log T \cdot d)）。SpikeGPT 和循环神经网络表现出较低的空间复杂度，这在资源有限的部署中是一个优势，尽管循环神经网络缺乏并行化支持。
Parallelization is crucial for scaling up training processes, and architectures like Transformer, Reformer, Linear Transformers, SpikeGPT, and AstroSNN support it, significantly reducing training time. Among these, Reformer excels in handling longer sequences due to its \(O(T \log T \cdot d)\) time complexity. SpikeGPT and Recurrent Neural Networks exhibit low space complexity, an advantage in resource-limited deployments, though RNNs lack parallelization support.

\begin{figure}[htbp]
    \centering
    \includegraphics[width=\linewidth]{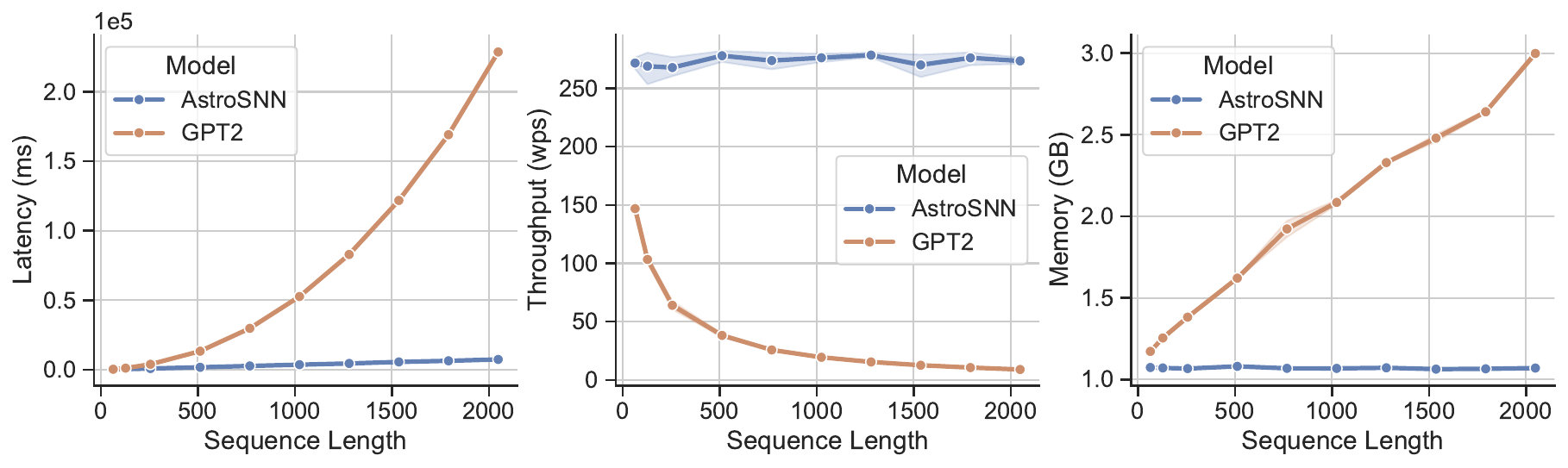}
    \caption{Comparative Performance of AstroSNN and GPT2 Models During Inference. (A) Latency across varying sequence lengths. (B) Throughput in words per second. (C) Memory Usage in gigabytes.}
    \label{fig:infer_cost}
\end{figure}

% 为了进一步说明AstroSNN的推理过程的优势, 我们在英特尔 13900K CPU 上进行了分析，比较了 AstroSNN 和 GPT2 共 24 层中 1.5B 模型的单层。结果如图~ref{fig:infer_cost}所示，显示了 AstroSNN 在不同序列长度下的延迟、吞吐量和内存使用方面的卓越性能。
To further illustrate the advantages of the inference process of AstroSNN, we conducted an analysis on an Intel 13900K CPU, comparing a single layer of the AstroSNN and GPT-2 models with a total of $24$ layers and a size of $1.5$ billion parameters. The results, as shown in Figure~\ref{fig:infer_cost}, demonstrate the outstanding performance of AstroSNN in terms of latency, throughput, and memory usage across different sequence lengths.

\begin{figure}[htbp]
    \centering
    \includegraphics[width=\textwidth]{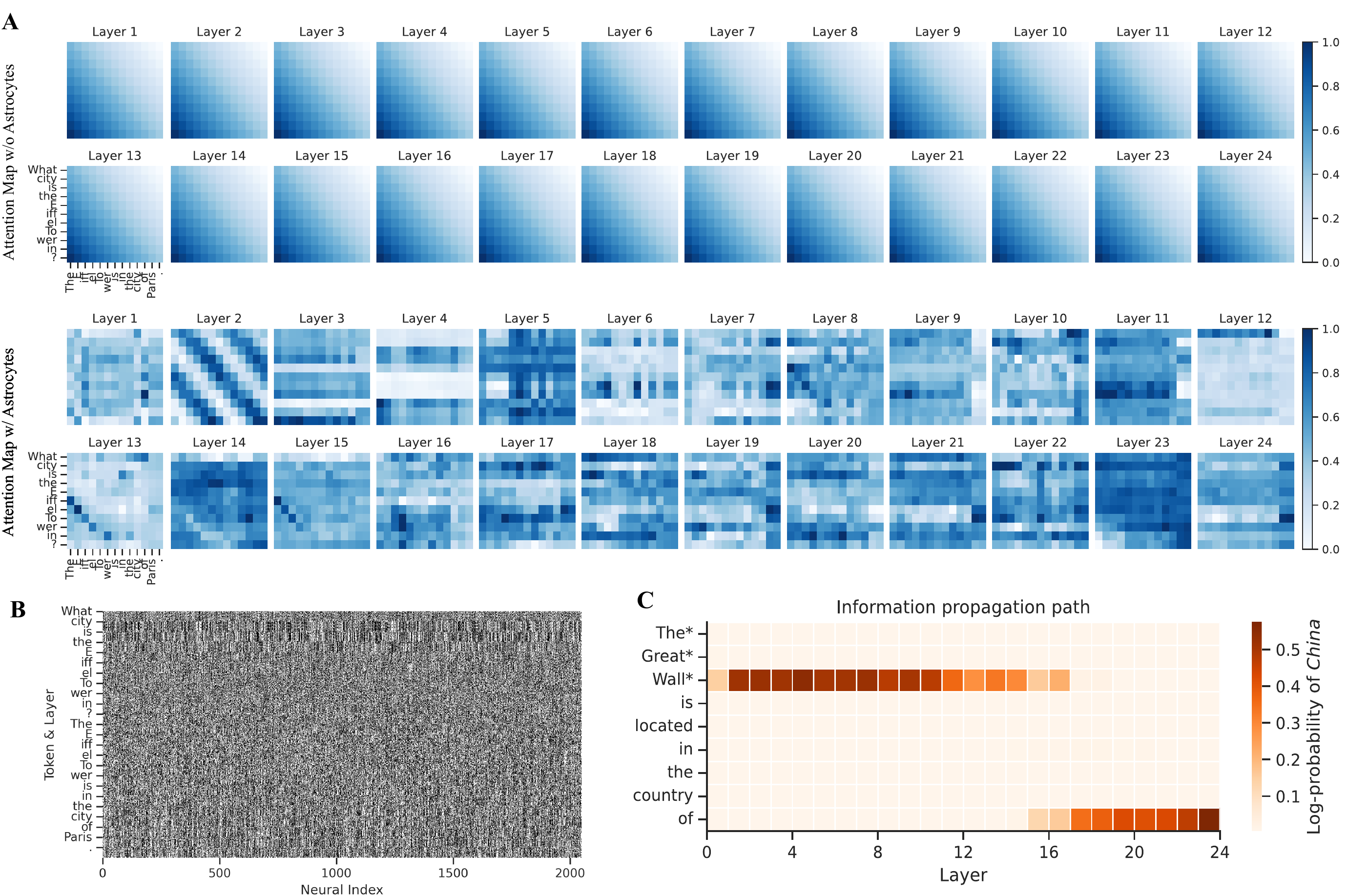}
    % A）AstroSNN 中的注意图显示了不同标记和层之间的不同注意，表明信息流丰富。(B) 不同层的神经元活动，纵轴代表标记，相邻标记表示不同层。(C)信息传播路径，揭示了AstroSNN如何保存和利用信息进行预测，突出了其在没有KV缓存的情况下保持信息相关性的能力。
    \caption{(A) Attention maps in AstroSNN show diverse attention across tokens and layers, indicating rich information flow. (B) Neuronal activity in different layers, with vertical axis representing tokens and adjacent marks indicating different layers. (C) Information propagation path, revealing how AstroSNN preserves and utilizes information for prediction, highlighting its capability to maintain information relevance without KV caches.}
    \label{fig:attention_maps}
\end{figure}

% 具体来说，与 GPT2 相比，AstroSNN 的延迟一直较低，在较长的序列长度上尤为明显。这种效率源于其相对于序列长度的线性复杂性，与传统转换器的二次复杂性形成鲜明对比。AstroSNN 即使在处理较长的序列时也能保持稳定的吞吐量，这对于处理长格式内容来说是一个显著优势。在内存消耗方面，与 GPT2 的急剧增加不同，AstroSNN 显示出与序列长度近乎恒定的曲线。这种内存效率归功于 AstroSNN 的递归计算策略，最大限度地减少了中间状态的存储。
Specifically, AstroSNN demonstrates consistently lower latency compared to GPT2, especially noticeable at longer sequence lengths. This efficiency stems from its linear complexity relative to sequence length, contrasting with the traditional Transformer's quadratic complexity. AstroSNN maintains stable throughput even for longer sequences, a significant advantage for processing long-form content. In terms of memory consumption, AstroSNN shows a near-constant profile regardless of sequence length, unlike GPT2's steep increase. This memory efficiency is attributed to AstroSNN's recursive computation strategy, minimizing the storage of intermediate states.

% \subsection{Visualization of AstroSNN} 

To demonstrate the dynamic interactions within the AstroSNN model, we visualize the attention matrices across different tokens, contrasting those with and without astrocytic integration (Figure~\ref{fig:attention_maps}). These visualizations highlight the enriched expressiveness of the astrocyte-enhanced attention mechanism compared to traditional LIF neurons.

% 为了展示 AstroSNN 模型中的动态交互作用，我们将不同标记的注意力矩阵可视化，并将有无星形胶质细胞整合的注意力矩阵进行对比（Figure~ref{fig:attention_maps}）。与传统的 LIF 神经元相比，这些可视化效果突出了星形胶质细胞增强注意力机制的丰富表现力。

% 在 AstroSNN 模型中，星形胶质细胞的增强使神经元能够更有效地进行交流，从而保留跨时间状态的信息。图{ref{fig:attention_maps}A显示了不同时间步的注意矩阵，揭示了神经元间细微的相互作用和强大的信息保持能力。图 B 侧重于同一序列中的神经元活动。

In the AstroSNN architecture, the integration of astrocytic mechanisms significantly enhances neuronal communication, ensuring effective preservation of information across temporal states. As depicted in Figure~\ref{fig:attention_maps} A, the attention matrices at varied timesteps exhibit sophisticated inter-neuronal interactions and a potent capacity for retaining information. Concurrently, Figure~\ref{fig:attention_maps} B concentrates on neuronal dynamics within identical sequences.
% In the AstroSNN model, astrocytic augmentation enables neurons to communicate more effectively, preserving information across temporal states. Figure~\ref{fig:attention_maps} A shows the attention matrices at various timesteps, revealing a nuanced inter-neuronal interaction and a robust information retention capacity. Figure~\ref{fig:attention_maps} B focuses on neuron activity within the same sequence.

% 我们使用 Meng 等人的可视化方法来评估模型保留和利用信息的能力。这一过程包括运行模型以记录层状态，用噪声干扰输入嵌入，然后在计算过程中恢复特定层状态。AstroSNN 模型在其隐藏状态中展示了高效的信息保留能力，无需依赖传统的 KV 缓存存储即可实现线性复杂度推理。这一特点使 AstroSNN 的早期层能够有效捕捉和传播关键句子元素，从而促进复杂语言结构的理解和生成。
Employing the visualization methodology proposed by Meng et al.~\cite{meng2022locating}, we evaluate the model's efficacy in sustaining and manipulating information. This entails executing the model to record layer states, introducing perturbations in input embeddings via noise, and subsequently reinstating specific layer states during the computational process. The AstroSNN model manifests a proficient retention of information within its hidden states, facilitating linear complexity inferences while obviating the need for conventional Key-Value (KV) cache storage. This attribute empowers the model’s initial layers to adeptly identify and convey pivotal sentence components, thereby enhancing its capability to comprehend and generate intricate linguistic constructs.
% Using the visualization method by Meng et al.~\cite{meng2022locating}, we assess the model's ability to retain and utilize information. This process involves running the model to record layer states, disrupting input embeddings with noise, and then restoring specific layer states during computation. The AstroSNN model demonstrates efficient information retention within its hidden states, enabling linear complexity inference without relying on traditional KV cache storage. This feature allows the early layers of AstroSNN to effectively capture and propagate key sentence elements, facilitating comprehension and generation of complex language structures.

\section{Discussion}

The development of the AstroSNN model represents a significant breakthrough in neuromorphic computing, seamlessly integrating biological insights with advanced computational techniques. This innovative approach addresses the limitations inherent in traditional LIF-based SNNs, enhancing both expressivity and the handling of complex temporal dependencies. Our extensive evaluations, including memory length and umbrella length tasks, demonstrate AstroSNN's superior performance in retaining and utilizing long-term dependencies, essential in cognitive tasks such as learning and memory. Furthermore, in the realm of Natural Language Generation (NLG), AstroSNN's scalability and efficiency open new horizons. Our experiments with large-scale language models showcase AstroSNN's capability to achieve competitive performance with significantly lower complexity than conventional Transformers. This efficiency is particularly evident in tasks requiring intricate linguistic understanding and generation, where AstroSNN exhibits capabilities akin to extensive language models while operating with a fraction of their computational load.

The incorporation of astrocytes in the AstroSNN model not only enhances the biological plausibility of SNNs but also introduces novel computational dynamics, making AstroSNN well-suited for applications in resource-constrained environments. This efficiency extends to potential use cases in edge computing and mobile devices, where computational resources are limited. The AstroSNN model opens new avenues in neuromorphic computing, particularly in designing systems that closely mimic biological neural processes. By capturing the essence of neuron-glial interactions, AstroSNN may pave the way for more advanced models that integrate other aspects of neural biology, potentially leading to breakthroughs in understanding brain function and developing more efficient computational models. However, the current implementation, though effective, is an abstraction of the complex dynamics in actual neuron-astrocyte interactions. Future research could explore more intricate models of glial cells, delve into different types of astrocytes, and their diverse roles in the brain. Additionally, extending the model to incorporate other forms of neural plasticity and adaptation could further enhance its capabilities. In conclusion, the astrocyte-modulated SNN introduced in this work not only challenges the current paradigms in neural network architectures but also provides a rich platform for further exploration in both computational neuroscience and artificial intelligence. Its success in various cognitive and computational tasks underscores the potential of biologically-inspired models in advancing AI, offering a glimpse into the future where the convergence of neuroscience and computing could lead to unprecedented technological advancements.

\section{Method}

\subsection{Experimental Setup}

% 我们的模型架构植根于当代的纯解码器方法（decoder-only approach~\cite{radford2019language}），这种方法常见于 GPT 等大规模语言模型中。在自回归任务中，模型会根据前面的标记预测序列中的下一个标记，而纯解码器范式因其简单有效而在该领域备受瞩目。通过专注于这一架构，我们充分利用了它在生成连贯且与上下文相关的文本序列方面的固有能力，使其成为我们的星胞修改的理想候选对象。
\paragraph{Architecture} The architecture of our model is rooted in the contemporary decoder-only approach~\cite{radford2019language}, commonly seen in large-scale language models such as GPT. The decoder-only paradigm has gained traction in the field due to its simplicity and effectiveness in autoregressive tasks where the model predicts the next token in a sequence based on the preceding ones. By focusing on this architecture, we leverage its inherent proficiency in generating coherent and contextually relevant text sequences, making it an ideal candidate for our astrocytic modifications.

% 在我们的设置中，我们用新颖的 AstroSNN 单元取代了 GPT 模型中的 Transformer 模块，从而将星形胶质细胞的功能无缝整合到已建立的神经框架中。我们的实现基于 gpt-neox，使其更容易扩展到更大的规模。
In our setup, we have replaced the Transformer module of the GPT model with our novel AM-SU unit, creating a seamless integration of astrocytic functions into the established neural framework. Our implementation is based on the gpt-neox~\cite{gpt-neox-library}, to make it easier to scale to larger sizes.

\paragraph{Tokenizer}
% 在我们的模型中，输入处理从标记化开始，使用 GPT-NeoX-20B tokenizer 将原始文本转换为神经网络可以理解的标记。这种标记化器善于处理各种语言和特殊字符，因此适用于各种语言数据集。在我们模型的初始层中，这些标记作为直接输入，为后续层奠定了基础。
In our model, the input processing begins with tokenization, where the \texttt
{GPT-NeoX-20B} tokenizer~\cite{black2204gpt} is employed to convert raw text into tokens that the neural network can understand. This tokenizer is adept at handling a wide array of languages and special characters, making it versatile for diverse linguistic datasets. In the initial layer of our model, these tokens serve as direct inputs, creating the foundation for the subsequent layers to build upon.

% 随着这一过程的展开，此后的每一层都将上一层的尖峰输出作为其输入，从而有效地形成了一个顺序结构。这种设计选择确保了信息在网络中的级联，其方式类似于刺激在生物神经网络中的流动，使每个连续层都能建立在输入标记的转换表征基础上。
As the process unfolds, each layer thereafter takes the spike outputs from the previous layer as its input, effectively forming a sequential structure. This design choice ensures that the information cascades through the network in a manner akin to the flow of stimuli through biological neural networks, allowing each successive layer to build upon the transformed representations of the input tokens.

\paragraph{Positional Encoding}
% 位置信息对于我们的模型理解序列中标记之间的顺序和关系至关重要。为了整合这些信息，我们采用了旋转位置编码（RoPE）技术，该技术以一种不受序列长度影响的方式将位置编码到注意力机制中。RoPE 对查询矢量和关键矢量进行旋转，可表示如下：
Positional information is crucial for our model to understand the order and relationship between tokens in a sequence. To integrate this information, we employ the Rotary Positional Encoding (RoPE)~\cite{su2021roformer} which encodes the position into the attention mechanism in a way that is invariant to sequence length. The RoPE applies a rotation to the query and key vectors, which can be represented as follows:

\begin{equation}
    \begin{aligned}
        \text{RoPE}(q_t, k_t, t) & = (q_t \cos(\omega_t) + k_t \sin(\omega_t),\ -q_t \sin(\omega_t) + k_t \cos(\omega_t)) \\
        \text{where}\ \omega_t   & = \text{position\_encoding}(t)
    \end{aligned}
    \label{eq:pos_embed}
\end{equation}

% 在方程~\ref{eq:pos_embed}中，\(q_i\)和\(k_i\)分别是查询向量和关键向量的第i个分量，\(\omega_i\)是导出的旋转频率 来自位置 \( i \) 的位置编码函数。 函数 \( \text{position\_encoding}(i) \) 为每个位置 \( i \) 生成唯一的编码。 旋转确保捕获相对位置，同时与绝对位置无关，从而保持模型的并行计算能力。
In Equation~\ref{eq:pos_embed}, \( q_t \) and \( k_t \) are the $t$-th components of the query and key vectors, respectively, and \( \omega_t \) is the rotation frequency derived from the positional encoding function for position \( t \). The function \( \text{position\_encoding}(t) \) generates a unique encoding for each position \( t \). The rotation ensures that the relative positioning is captured while being agnostic to the absolute positions, thus maintaining the model's capacity for parallel computation.

% RoPE 的特别优势在于它能够允许模型保留标记的相对顺序并支持模型对输入的顺序性质的理解，而无需牺牲计算效率。 这种编码方法在不同的序列长度上都具有鲁棒性和一致性，使其成为我们的 AstroSNN 模型的理想选择，因为输入序列的动态范围和上下文可能非常广泛。
The RoPE is particularly advantageous for its ability to allow the model to preserve the relative order of the tokens and to support the model's understanding of the sequential nature of the input without sacrificing computational efficiency. This encoding method is robust and consistent across varying sequence lengths, making it an ideal choice for our AstroSNN model where the dynamic range and context of the input sequence can be quite extensive.

\paragraph{Time Constants of Neurons and Astrocytes}

% 神经元和星形胶质细胞对突触输入的动态响应在很大程度上受到各自膜时间常数的影响。神经元需要捕捉快速变化的信号，其膜时间常数设定为 2，这有利于快速整合输入并处理快速的时间变化。神经元的静息电位v_th被设置为0， 膜电阻R被设置为1.
The dynamism of neural and astrocyte response to synaptic inputs is heavily influenced by their respective membrane time constants. Neurons, which need to capture swiftly varying signals, have a membrane time constant $\tau_n$ set to $2$, facilitating quick integration of inputs and enabling the processing of rapid temporal changes. The resting potential $v_{th}$ is set to $0$ and the membrane resistance $R$ is set to $1$.

% 星形胶质细胞在较长期的突触变化中扮演着重要角色，因此我们使用一系列时间常数对其进行建模，以反映其不同的时间整合能力。这种多样性在我们的多头注意机制中表现得尤为明显，在该机制中，每个头都被分配了一个独特的时间常数，从而使网络能够跨越连续的时间依赖关系。在实验中, $\tau_a$ 被设置为[32, 512] 之间按照指数分布采样的值, 采样数量是模型的头数, 在实验中被设置为$8$。
Astrocytes, distinguished by their role in longer-term synaptic changes, are modeled with a range of time constants to reflect their varied temporal integration capabilities. This diversity is particularly manifest in our multi-headed attention mechanism, where each head is allocated a unique time constant, allowing the network to span a continuum of temporal dependencies.

Astrocytes play a crucial role in long-term synaptic changes, hence we model them with a range of time constants to reflect their varying capacities for temporal integration. This diversity is particularly evident in our multi-head attention mechanism, where each head is allocated a unique time constant, allowing the network to encompass a continuum of temporal dependencies. In our experiments, the time constant for astrocytes, $\tau_a$, is set to values exponentially distributed between $32$ and $512$, with the number of samples equaling the number of heads in the model, which is set to 8 in our case.

% 由此产生的保留掩码使每个头能够专注于不同的时间窗口，一些头捕捉到更直接的互动，而另一些头则在更长的时间跨度内进行整合。这种细致入微的时间编码方法使 AstroSNN 模型能够在短期反应和长期背景整合之间保持精密的平衡，使其能够理解具有类似生物神经处理的复杂性和深度的序列。
The resulting $\tau_a$ enable each head to focus on a distinct temporal window, with some heads capturing more immediate interactions and others integrating over a longer time span. This nuanced approach to temporal encoding allows the AstroSNN model to maintain a sophisticated balance between short-term reactivity and long-term contextual integration, endowing it with the ability to understand sequences with complexity and depth akin to biological neural processing.

\paragraph{Hyperparameters}

\begin{table}[ht]
    \centering
    \caption{Hyperparameter Settings for Different Models.}
    \begin{tabular}{l|cccc}
        \hline

        \textbf{Hyperparameters}    & \textbf{\tabincell{c}{AstroSNN                                                                                                                                                     \\ Small}} & \textbf{\tabincell{c}{AstroSNN \\ Base}} & \textbf{\tabincell{c}{AstroSNN \\ Large}} & \textbf{\tabincell{c}{AstroSNN \\ 1B5}} \\
        \hline
        Layer                       & $12$                                                                                                                                       & $16$       & $20$       & 24          \\
        Dim                         & $512$                                                                                                                                      & $1024$     & $1536$     & $2048$      \\
        Parameters                  & $88.186M$                                                                                                                                  & $306.412M$ & $732.678M$ & $1444.984M$ \\
        \hline
        Learning Rate               & \multicolumn{4}{c}{$2.5 \times 10^{-4}$}                                                                                                                                           \\
        LR Scheduler                & \multicolumn{4}{c}{cosine}                                                                                                                                                         \\
        Minimum Learning Rate       & \multicolumn{4}{c}{$6 \times 10^{-5}$}                                                                                                                                             \\
        Warm-up Steps               & \multicolumn{4}{c}{$1000$}                                                                                                                                                         \\
        Tokens per Batch            & $320k$                                                                                                                                     & $160k$     & $80k$      & $64k$       \\
        Adam \(\beta\)              & \multicolumn{4}{c}{($0.9$, $0.98$)}                                                                                                                                                \\
        Training Steps              & \multicolumn{4}{c}{$600000$}                                                                                                                                                       \\
        Gradient Clipping           & \multicolumn{4}{c}{$1.0$}                                                                                                                                                          \\
        Weight Decay                & \multicolumn{4}{c}{$0.1$}                                                                                                                                                          \\
        \hline
        Surrogate Gradient Function & \multicolumn{4}{c}{$g^\prime(x) = \alpha^2 \left|x\right| + \alpha, \quad \text{if} \left|x\right| \leq 1 / \alpha, \text{else} \quad 0$ }                                         \\
        \hline
    \end{tabular}
    \label{tab:hyperparameters}
\end{table}

% 为了满足各种计算资源和性能目标，我们配置了多个版本的 AstroSNN 模型，每个版本都有一套量身定制的超参数。最小的配置 AstroSNN-Small 包含 12 层，维度为 512，共约 8,818.6 万个参数。AstroSNN-Base 和 AstroSNN-Large 模型在此基础上进一步扩展，分别有 16 层和 20 层，维度增加后，参数数分别为 3.06412 亿和 7.32678 亿。最大的配置 AstroSNN-1B5 拥有 24 层，维度为 2048，最终得到近 14.45 亿个参数。
To cater to a variety of computational resources and performance targets, we have configured multiple versions of the AstroSNN model, each with a tailored set of hyperparameters. The smallest configuration, AstroSNN-Small, consists of 12 layers with a dimensionality of 512, amounting to approximately 88.186 million parameters. The AstroSNN-Base and AstroSNN-Large models extend this further with 16 and 20 layers, and increased dimensionality leading to 306.412 million and 732.678 million parameters, respectively. The largest configuration, AstroSNN-1B5, boasts 24 layers with a dimensionality of 2048, culminating in nearly 1.445 billion parameters.

% 所有模型都共享一个统一的学习率（2.50乘以10^{-4}），在整个训练过程中使用余弦学习率调度器来调整学习率。为防止学习率降为零，设置了最低学习率（6.00 （乘以 10^{-5}））。在所有配置中，热身期都被标准化为 1000 步，为模型的收敛做好准备。以代币为单位的批量大小与模型大小成反比变化，最小的模型为 320k，最大的模型为 64k，以确保在 GPU 内存限制范围内容纳较大的模型，因为这些模型更耗费内存。
All models share a uniform learning rate of \(2.50 \times 10^{-4}\), with a cosine learning rate scheduler~\cite{loshchilov2016sgdr} to adapt the learning rate throughout training. A minimum learning rate of \(6.00 \times 10^{-5}\) is set to prevent the rate from decreasing to zero. The warm-up period is standardized at 1000 steps across all configurations, preparing the models for convergence. The batch size in tokens varies inversely with model size, ranging from 320k for the smallest model down to 64k for the largest, ensuring that larger models, which are more memory-intensive, are accommodated within GPU memory constraints.

% 优化采用亚当算法，其值分别为 0.9 和 0.98。这一选择平衡了对动力的需求和对新梯度的响应。计划总共进行 600,000 步训练，为模型吸收语言的复杂性提供一个全面的学习期。梯度削波的阈值为 1，以保持训练的稳定性。最后，采用 0.1 的权重衰减进行正则化，以降低过度拟合的风险。如Equation~\ref{eq:output}所示, 由于$g(x)$不可微分, 因此需要使用代理梯度函数建立近似的反向传播的通路, 我们使用的代理梯度函数如Table~\ref{tab:hyperparameters}所示。
The optimization is carried out using the Adam~\cite{kingma2014adam} with \(\beta\) values of $0.9$ and $0.98$. This choice balances the need for momentum with the responsiveness to new gradients. A total of $600,000$ training steps are planned, providing a comprehensive learning period for the model to assimilate the intricacies of the language. Gradient clipping is employed with a threshold of 1 to maintain training stability. Lastly, a weight decay of 0.1 is applied for regularization, mitigating the risk of overfitting. As shown in Equation~\ref{eq:output}, since $g(x)$ is not differentiable, it is necessary to use the surrogate gradient function~\cite{bohte2011error} to establish an approximate backpropagation pathway, and we use the surrogate gradient function as shown in Table~\ref{tab:hyperparameters}.

% 我们使用朴素的PyTorch代码实现了AstroSNN, 在The Pile上的预训练使用了4张NVIDIA A100 40G GPU, 并使用了张量并行. 模型的预训练共使用了38.4B个标记.
We implemented AstroSNN using plain PyTorch code, and pre-training on The Pile dataset used 4 NVIDIA A100 40G GPUs with tensor parallelism. A total of $38.4$B tokens were used for the pre-training of the model.

\subsection{Data Availability}

% 本研究中使用的数据集跨越了各种领域和任务，确保 AstroSNN 模型能够在一整套挑战中得到评估。在评估模型的记忆和时间处理能力的任务中，我们使用了 DeepMind 的 bsuite 数据集中的 "记忆长度 "和 "伞状长度 "环境。我们使用Jax重构的环境, 以实现高效的训练和评估, 上述环境可以在https://github.com/RobertTLange/gymnax/tree/main/gymnax/environments/bsuite找到.
The datasets employed in this study span a variety of domains and tasks, ensuring that the AstroSNN model is evaluated across a comprehensive suite of challenges. For tasks assessing the model's memory and temporal processing capabilities, we utilize the Memory Length and Umbrella Length datasets from DeepMind's bsuite collection. We use a re-implemented environment using Jax for efficient training and evaluation, which can be found at \footnote{Bsuite is available at \url{https://github.com/RobertTLange/gymnax/tree/main/gymnax/environments/bsuite}}.

% 在 NLP 领域，我们利用了 enwiki8 (\url{https://huggingface.co/datasets/enwik8}) 和 wikitext-103 (\url{https://huggingface.co/datasets/wikitext}) 数据集。enwiki8 数据集是英文维基百科文本的大型语料库，用于深度学习模型的无监督学习。Wikitext-103是从维基百科上经过验证的优秀文章和精选文章中提取的超过1亿个词块的集合，通常用于语言模型的基准测试。
For NLP tasks, we leverage the enwiki8 \footnote{Enwiki8 dataset is available at \url{https://huggingface.co/datasets/enwik8}} and wikitext-103 \footnote{Wikitext-103 dataset is available at \url{https://blog.einstein.ai/the-wikitext-long-term-dependency-language-modeling-dataset/}} datasets. The enwiki8 dataset is a large corpus of English Wikipedia text, used for unsupervised learning of deep learning models. Wikitext-103 is a collection of over 100 million tokens extracted from the set of verified Good and Featured articles on Wikipedia and is commonly used for benchmarking language models.

% 为了进行零点学习实验，我们策划了一系列不同的数据集，包括 PIQA、SciQ、Winograd、COPA、BoolQ、hellaswag 和 logiqa。这些数据集提供了从常识推理、科学事实理解到语言推理等各种任务，为评估模型的零点学习能力提供了严格的测试平台。
For our zero-shot learning experiments, we have curated a diverse array of datasets including PIQA \footnote{PIQA dataset is available at \url{https://yonatanbisk.com/piqa/}}, SciQ \footnote{SciQ dataset is available at \url{https://allenai.org/data/sciq/}}, Winograd \footnote{WinoGrande dataset is available at \url{https://winogrande.allenai.org/}}, COPA \footnote{COPA dataset is available at \url{https://people.ict.usc.edu/~gordon/copa.html}}, BoolQ \footnote{BoolQ dataset is available at \url{https://github.com/google-research-datasets/boolean-questions}}, hellaswag \footnote{Hellaswag dataset is available at \url{https://rowanzellers.com/hellaswag/}}, and logiqa \footnote{Logiqa dataset is available at \url{https://github.com/lgw863/LogiQA-dataset}}. These datasets present tasks that range from commonsense reasoning and scientific fact understanding to linguistic inference, providing a rigorous test bed for evaluating the zero-shot learning abilities of the model.

% 在进行这些实验之前，我们在 The Pile 数据集上对模型进行了预训练，该数据集可在 \url{https://pile.eleuther.ai/}} 上获取，它是一个大规模的数据集，由各种来源汇编而成，提供了丰富多样的语言建模基准。该数据集旨在挑战各种语言现象和学科专业知识的模型。对于对话模型的微调, 我们使用了MOSS微调数据集. 本文提供了源代码。
Prior to these experiments, the model was pre-trained on The Pile \footnote{The Pile dataset is available at \url{https://pile.eleuther.ai/}}, a large-scale dataset compiled from various sources to provide a rich and diverse language modeling benchmark. The Pile is designed to challenge models with a wide range of linguistic phenomena and subject matter expertise. For the fine-tuning of the dialog model, we used the MOSS fine-tuning dataset \footnote{MOSS fine-tuning dataset is available at \url{https://huggingface.co/datasets/fnlp/moss-003-sft-data}}. Source code are provided with this paper.

\section*{Acknowledgement}
This work was supported by the National Natural Science Foundation of China (Grant No. 62372453).

\bibliography{refs}
\bibliographystyle{unsrt}

\section*{Supplementary Material}
\subsection*{Examples of Dialogues Generated by AstroSNN}

\label{sec:examples}

\begin{dialoguebox}[General Knowledge \#1]\label{test1}
    \textbf{Human:} Can you explain to me what astrocytes are? \\
    \textbf{Mind Talk:} Certainly! Astrocytes are specialized cells found in the brain that are responsible for the production of neurotransmitters and other molecules that are involved in the development of the brain. They are also involved in the regulation of blood flow and oxygen delivery to the brain. Astrocytes are found in the brain and spinal cord, and they are believed to play a crucial role in the development of the brain.
\end{dialoguebox}

\begin{dialoguebox}[General Knowledge \#2]
    \textbf{Human:} What causes the seasons to change on Earth?\\
    \textbf{Mind Talk:} The seasons change on Earth due to various factors such as climate change, natural disasters, and human activities.
\end{dialoguebox}

\begin{dialoguebox}[Technical Explanations: \#1]
    \textbf{Human:} Could you describe the theory of relativity in simple terms?\\
    \textbf{Mind Talk:} Certainly! The theory of relativity is a scientific theory that explains how the laws of physics work at the atomic and subatomic level. It is based on the idea that the speed of light, which is the speed of light in vacuum, is not a constant but is instead a relative speed. This means that the speed of light can change depending on the observer's position in the universe. The theory of relativity also explains how time and space are related, as well as how gravity works. It is a very important and influential theory in modern physics.
\end{dialoguebox}

\begin{dialoguebox}[Code Generation \#1]
    \textbf{Human:} Write me a quick sort algorithm in python.\\
    \textbf{Mind Talk:}  Sure, here's a quick sort algorithm in Python:

    \begin{verbatim}
'''
def quick_sort(arr):
    if len(arr) <= 1:
        return arr
    else:
        pivot = arr[0]
        less = [x for x in arr[1:] if x <= pivot]
        greater = [x for x in arr[1:] if x > pivot]
        return quick_sort(less) + [pivot] + quick_sort(greater)
'''
\end{verbatim}

    This algorithm first checks if the length of the array is less than or equal to 1. If so, it returns the array as it is already sorted. Otherwise, it selects the first element of the array as the pivot and creates two sub-arrays: one containing all elements less than or equal to the pivot, and one containing all elements greater than or equal to the pivot. It then recursively calls the quick\_sort function on the sub-arrays and concatenates the sorted sub-arrays with the pivot element in between.

    Here's an example usage of the function:

    \begin{verbatim}
'''
arr = [5, 3, 8, 4, 2, 7, 1, 6]
sorted_arr = quick_sort(arr)
print(sorted_arr)
'''
\end{verbatim}

    This will output: [1, 2, 3, 4, 5, 6, 7, 8]
\end{dialoguebox}

\begin{dialoguebox}[Philosophical Questions \#2]
    \textbf{Human:} Is there such a thing as free will?\\
    \textbf{Mind Talk:} As an AI language model, I don't have personal beliefs or opinions. However, I can provide you with some information on the topic of free will.

    Free will is a philosophical concept that refers to the ability of an individual to make choices and decisions without being explicitly programmed to do so. It is often referred to as the "control over one's own destiny" or "control of one's own life."

    While some people believe that free will is a fundamental aspect of the human condition, others argue that it is not. Some argue that free will is a product of biological processes that are not fully understood, while others believe that it is a result of complex interactions between the brain, body, and environment.

    In any case, it is important to note that free will is a complex and multifaceted concept that is still being explored by scientists and philosophers. While some people may argue that free will is a uniquely human trait, others argue that it is a product of complex biological processes that are not fully understood.
\end{dialoguebox}

\begin{dialoguebox}[Technical Explanations: \#2]
    \textbf{Human:} How does a blockchain work?\\
    \textbf{Mind Talk:} A blockchain is a decentralized, digital ledger that records transactions in a secure and transparent manner. It is made up of a series of blocks, each containing a set of transactions that are verified and added to the blockchain. Each block contains a unique code that is used to validate the transaction and add it to the blockchain. Once a block is added to the blockchain, it cannot be altered or deleted, making it a tamper-proof and secure way to record transactions
\end{dialoguebox}

\begin{dialoguebox}[Creative and Imaginative \#1]
    \textbf{Human:} If you could design a city of the future, what features would it have?\\
    \textbf{Mind Talk:} As an AI language model, I don't have personal preferences or emotions, but I can provide some ideas for a futuristic city.

    One of the most important features of a futuristic city would be the integration of technology. The city would be designed with the latest and most advanced technology, such as artificial intelligence, biotechnology, and nanotechnology. This would allow for a more efficient and sustainable way of living, with advanced transportation systems, energy sources, and waste management systems.

    Another important feature would be the use of advanced communication technology. The city would have a highly advanced communication system that would allow for real-time communication between people, businesses, and other organizations. This would be especially important in a world where communication is limited, as it would allow for more efficient and effective decision-making.

    The city would also have a highly advanced infrastructure, including advanced transportation systems, energy sources, and waste management systems. This would allow for a more sustainable and environmentally friendly way of living, with a focus on reducing carbon emissions and promoting sustainability.

    Finally, the city would have a highly advanced healthcare system, with advanced medical technology and treatments. This would allow for a more efficient and effective way of treating illnesses and diseases, as well as improving the overall quality of life for people.
\end{dialoguebox}

\begin{dialoguebox}[Cultural and Entertainment \#1]
    \textbf{Human:} Can you recommend a good movie for a movie night with friends?\\
    \textbf{Mind Talk:} Sure, I can recommend a good movie for a movie night with friends.There are so many great movies to choose from! Here are a few popular movies that I would recommend for a movie night with friends:

    1. The Shawshank Redemption (1994) - A classic drama that tells the story of a man who is wrongfully convicted of murder and sent to prison.

    2. The Godfather (1972) - A classic crime drama that tells the story of the Corleone family and their involvement in organized crime.

    3. The Dark Knight (2008) - A superhero movie that explores the dark side of the Batman character and his impact on Gotham City.

    4. The Matrix (1999) - A sci-fi action movie that explores the nature of reality and the power of the human mind.

    5. The Green Knight (2001) - A fantasy adventure movie that follows a knight on a quest to save a princess from a dragon.

    6. The Silence of the Lambs (1991) - A psychological thriller that explores the nature of evil and the consequences of our actions.

    7. The Princess Bride (1987) - A romantic fantasy movie that follows the love story of Princess Diana and Prince William.

    These are just a few examples, but there are many other great movies out there to choose from!
\end{dialoguebox}

\begin{dialoguebox}[Practical Advice \#1]
    \textbf{Human:} What are some effective study techniques for exams?\\
    \textbf{Mind Talk:} There are many effective study techniques for exams, including:

    1. Active learning: Engage with the material actively by taking notes, summarizing key points, and explaining concepts to others.

    2. Spaced repetition: Review the material at increasing intervals to reinforce learning and improve retention.

    3. Practice testing: Test yourself on the material to identify areas of weakness and areas of strength.

    4. Mnemonics: Use mnemonic devices to help remember information.

    5. Visual aids: Use diagrams, charts, and other visual aids to help you understand and remember complex information.

    6. Teach someone else: Teach the material to someone else to help reinforce your understanding and identify areas where you need to improve.

    7. Take breaks: Take regular breaks to avoid burnout and maintain focus and concentration.

    8. Get enough sleep: Get enough sleep to help your brain function at its best and improve memory retention.

    9. Stay organized: Keep your notes and study materials organized to avoid wasting time searching for information.

    10. Seek help: If you are struggling with a particular topic, seek help from a teacher, tutor, or study group.
\end{dialoguebox}

\end{document}